\documentclass[10pt,twocolumn,letterpaper]{article}

\usepackage{cvpr}              

\usepackage{graphicx}
\usepackage{amsmath}

\DeclareMathOperator*{\argmin}{arg\,min}
\usepackage{amssymb}
\usepackage{booktabs}
\usepackage{url}

\usepackage[pagebackref,breaklinks,colorlinks]{hyperref}

\usepackage[capitalize]{cleveref}
\crefname{section}{Sec.}{Secs.}
\Crefname{section}{Section}{Sections}
\Crefname{table}{Table}{Tables}
\crefname{table}{Tab.}{Tabs.}

\def\cvprPaperID{60} 
\def\confName{CVPR}
\def\confYear{2022}

\begin{document}

\title{Manifold Regularization for Memory-Efficient Training of Deep Neural Networks}

\author{Shadi Sartipi\\
University of Rochester\\
500 Joseph C.~Wilson Blvd.~Rochester, NY 14627\\
{\tt\small SSartipi@ur.rochester.edu}
\and
Edgar A.~Bernal\\
FLXAI\\
44 Elton St.~Rochester, NY 14607\\
{\tt\small Edgar.Bernal@flxai.com}
}
\maketitle

\begin{abstract}
One of the prevailing trends in the machine- and deep-learning community is to gravitate towards the use of increasingly larger models in order to keep pushing the state-of-the-art performance envelope. This tendency makes access to the associated technologies more difficult for the average practitioner and runs contrary to the desire to democratize knowledge production in the field. In this paper, we propose a framework for achieving improved memory efficiency in the process of learning traditional neural networks by leveraging inductive-bias-driven network design principles and layer-wise manifold-oriented regularization objectives. Use of the framework results in improved absolute performance and empirical generalization error relative to traditional learning techniques. We provide empirical validation of the framework, including qualitative and quantitative evidence of its effectiveness on two standard image datasets, namely CIFAR-10 and CIFAR-100. The proposed framework can be seamlessly combined with existing network compression methods for further memory savings.
\end{abstract}

\section{Introduction}
\label{sec:intro}

As deep neural networks continue to push the state-of-the-art in areas such as computer vision and natural language processing (NLP), their complexity and size seem to inevitably grow larger. Four of the current top five performers on the ImageNet leaderboard~\cite{ImageNetLeaderboard} have over 2 billion parameters, namely, CoCa~\cite{Yu2022CoCaCC}, ModelSoups BASIC-L~\cite{pmlr-v162-wortsman22a}, PaLI~\cite{51651} and CoAtNet-7~\cite{Dai2021CoAtNetMC}, with the smallest of the five (ModelSoups ViT-G14~\cite{pmlr-v162-wortsman22a}) having 1.84 billion parameters. A 22-billion parameter vision transformer, termed ViT-22B, was recently proposed~\cite{https://doi.org/10.48550/arxiv.2302.05442} and was shown to match or improve upon state of the art across a range of vision tasks. Networks that excel at NLP tasks are often larger. Prior work has shown that empirical performance of language models has a power-law relationship with the number of parameters (model size), dataset size and computational budget~\cite{Kaplan2020ScalingLF}. Examples of top-performing NLP networks include GPT-3/ChatGTP~\cite{NEURIPS2020_1457c0d6}, the Switch Transformer Architecture~\cite{fedus2021switch}, and PaLM~\cite{Chowdhery2022PaLMSL}, with 175 billion, 1.6 trillion, and 540 billion parameters, respectively. Training GPT-3 alone is said to have cost \$12 million~\cite{GPT3cost} and incurred an estimated 78,000 of CO$_2$ emissions~\cite{GPT3emissions}. OpenAI research has estimated that the doubling in compute power cadence has gone from two years before 2012 to 3.4 months since then~\cite{computedoubling}.

The often unrealistic computational requirements imposed by deep learning models remain in stark contrast with the long-standing goal of equipping edge and IoT devices~\cite{10.1145/3131672.3131675} including smartphones, wearables~\cite{Bernal2018DeepTM}, appliances, autonomous vehicles~\cite{9478090}, and even satellites~\cite{https://doi.org/10.48550/arxiv.2001.10362} with machine- and deep-learning capabilities. This, in the hopes of bringing the compute power as close to the data sources and end users as possible~\cite{8976180}. Given the inherent computational limitations of edge devices, neural network compression has long been an area of intense research within the community~\cite{MARINO2023152,DBLP:journals/corr/abs-2010-03954}. Today, it is a particularly relevant field in light of the prohibitive requirements to train and even deploy state-of-the-art models which are often beyond the means of the average business, let alone individual practitioners. Traditional techniques for network compression include weight sharing, pruning, tensor decomposition, knowledge distillation, and quantization~\cite{DBLP:journals/corr/abs-2010-03954}. Unfortunately, most of the existing network compression techniques are \textit{post-hoc} in nature, meaning that they require a large model to be trained first before compression can be effected. This \textit{modus operandi} partially offsets the advantages brought about by the memory-saving nature of the techniques, as it limits their impact to the deployment stage. In this paper, we propose a framework to enable memory savings throughout the full model development life cycle, including the training stage, thus filling a void in the current literature. 

Model size is a significant contributor to consumed GPU memory, not only because the larger the model, the larger the number of learnable parameters that needs to be stored, but also the number of gradient values that needs to be tracked throughout the training stage. Another factor that drives memory usage during learning is the data itself. Assuming that the dataset size and dimensionality are fixed (i.e., because they are determined by the task), memory constraints associated with data can be ameliorated by implementing stochastic gradient descent (SGD) techniques~\cite{Bottou2004} and its variants, which compute a gradient estimate from a small number of samples termed a mini-batch. It has been shown that mini-batch size is one of the main contributors to memory usage during model training~\cite{10.1145/3368089.3417050}. While using smaller mini-batches does indeed reduce memory consumption, it has been shown that the variance in the gradient estimate increases as the mini-batch size decreases~\cite{https://doi.org/10.48550/arxiv.2004.13146}, which can lead to undesired convergence behavior~\cite{article}.

In this paper, we propose a new paradigm for achieving memory efficiency that addresses two of the main aspects that drive GPU memory consumption during the training process of a deep learning model, namely the number of learnable model parameters and the mini-batch size~\cite{10.1145/3368089.3417050}. To our knowledge, this work is the first of its kind to address both issues simultaneously. The motivation for the proposed framework stems from the insight that the underlying dimensionality of the data at hand is one of the main drivers of the complexity of the learning process~\cite{NIPS2010_8a1e808b,Narayanan2009OnTS}, and that traditional network architectures can therefore be compressed significantly \textit{a priori}. In short, our method enables the hierarchical feature learning process in deep neural networks to mimic a sequential manifold learning process by enforcing geodesic-distance-preserving objectives~\cite{8953348} on the intermediate representation spaces. This results in learning processes that are robust across a wide range of model and mini-batch sizes, and networks with improved inductive bias and generalization capabilities, thus enabling significant memory savings during the training process.

The contributions of this paper are as follows:

\vspace{-0.3cm}
\begin{itemize}
	\setlength\itemsep{-0.15cm}
	\item a framework to achieve memory-efficient development of deep neural networks throughout the model's full life cycle, i.e., not limited to \textit{post-hoc} implementation as most existing network compression techniques;
	\item a distance-preserving regularization mechanism operating on the intermediate network layers of the network that enables the use of extremely small mini-batch sizes; 
	\item an inductive-bias-driven network design principle that enables network compression from inception through training and deployment; and 
        \item experimental verification of the effectiveness of the proposed framework with regards to both absolute performance and empirical generalization error, as well as GPU memory usage.
\end{itemize}\vspace{-0.3cm}

\section{Related Work}
\subsection{Intrinsic Dimensionality and the Complexity of a Learning Task} It has long been acknowledged that data can usually be represented with a number of free parameters that is smaller than its ambient dimensionality~\cite{1054365}. This is referred to as \textit{intrinsic dimensionality}, and a number of algorithms for estimating it have been proposed~\cite{4766873,1039212,1671801,DBLP:conf/iclr/PopeZAGG21,8953348,osti_10344297}. Further, connections have been found between the concept of intrinsic dimensionality and the complexity of learning tasks. For instance, it has been shown that learning compact representations of data requires a number of samples that grows exponentially with the intrinsic dimensionality of the data~\cite{NIPS2010_8a1e808b}; similarly, learning a decision boundary between two classes requires a number of samples that grows exponentially with the intrinsic dimensionality of the space in which the classes lie~\cite{Narayanan2009OnTS}; lastly, datasets with low intrinsic dimensionality have been found to be easier to learn with deep neural networks, and the resulting models are better at generalizing between training and test data~\cite{DBLP:conf/iclr/PopeZAGG21}. 
\subsection{Manifold Learning} Closely related to the concept of intrinsic dimensionality is the construct of a manifold. Manifold learning involves the construction of a non-linear, dimensionality reducing mapping between the ambient space of the raw data and the low-dimensional manifold on which the data resides~\cite{NIPS2010_8a1e808b}. Examples of manifold learning methods include Locally Linear Embedding~\cite{doi:10.1126/science.290.5500.2323}, ISOMAP~\cite{doi:10.1126/science.290.5500.2319}, Laplacian Eigenmaps~\cite{6789755}, and Hessian Eigenmaps~\cite{doi:10.1073/pnas.1031596100}]. In the context of supervised deep learning, the concept of manifold is relevant because the task of the neural network is to find a coordinate representation of the data manifold such that the classes are linearly separable by hyperplanes~\cite{NIPS2017_0ebcc77d}. It has been estimated that the complexity of that task grows exponentially with the dimensionality of that manifold and polinomially with its curvature~\cite{NIPS2010_8a1e808b}. Leveraging the manifold structure of data has lead to contributions in adversarial robustness~\cite{Jin2020ManifoldRF}, multimodal fusion~\cite{Nguyen2021.01.28.428715}, robustness to noise~\cite{Tomar2014ManifoldRD} and semi-supervised learning~\cite{JMLR:v7:belkin06a}.

\section{Proposed Framework}

\subsection{Framework Description}
The proposed framework addresses two of the main contributors to memory consumption in the training process of a deep neural network, namely number of model parameters and mini-batch size. The operating principle behind our framework stems from the recognition that natural data lies on a low-dimensional manifold embedded in the higher-dimensional ambient space. From that standpoint, we posit that a neural network could achieve significant parameter economy by effecting bottlenecks in the data path with widths in the vicinity of the size of the intrinsic dimensionality of the data. However, it has been shown that narrow networks tend to showcase larger gradient estimate variance~\cite{ghosh2023how} and are in general harder to optimize~\cite{DBLP:journals/corr/abs-2010-00885}, repercussions that will be compounded by using small mini-batch sizes in an effort to further achieve memory savings. And, as is well known, in the presence of large gradient variance, the estimates of the network parameters will bounce around the target minima~\cite{article}. In order to counteract these undesired effects, we recognize that the task of training a neural network is equivalent to learning a sequence of coordinate transformations that start from the data representation in the ambient space, and end in a space whose coordinate representation is a lower-dimensional manifold that facilitates the performance of the task at hand as determined by the objective function~\cite{NIPS2017_0ebcc77d}. Further, we note that the dimensionality of the data representation computed by the network corresponds to the size (i.e., width) of its bottleneck layer~\cite{NIPS2017_0ebcc77d}. In view of these observations, we propose to enforce multi-dimensional scaling principles~\cite{kruskal1664}, more specifically, distance-preserving objectives~\cite{8953348} in an attempt to encourage the formation of manifold structures across the multiplicity of dimensionality reduction stages that take place as the data traverses the network. 

The use of a network having a bottleneck width commensurate with the intrinsic dimensionality of the data is aimed at leveraging inductive bias~\cite{DBLP:conf/iclr/CohenS17,Ulyanov_2018_CVPR} which inherently constraints the solution space via the architectural choice itself. The layer-wise regularization which is integral to our framework is related to the use of priors in inverse problems~\cite{NIPS2009_3707} and data pausity scenarios~\cite{Bernal_2021_CVPR}, except that, in our case, the prior is enforced in the intermediate feature spaces, as opposed to the ambient space itself, and the data scarcity is not due to lack of available data but to the memory constraints being imposed. The features thusly learned, although based on limited observations of the training set, showcase stability facilitated by the preserved local attributes of the intermediate activation spaces, not unlike features from previous works aimed at building robust data representations in an unsupervised manner~\cite{5539957}. As will become clear later, this results in networks that showcase improved generalization capabilities, as measured by the difference between the performance on the training and test sets.



\subsection{Framework Formulation}
\begin{figure}[t]
  \centering
   \includegraphics[width=1.0\columnwidth]{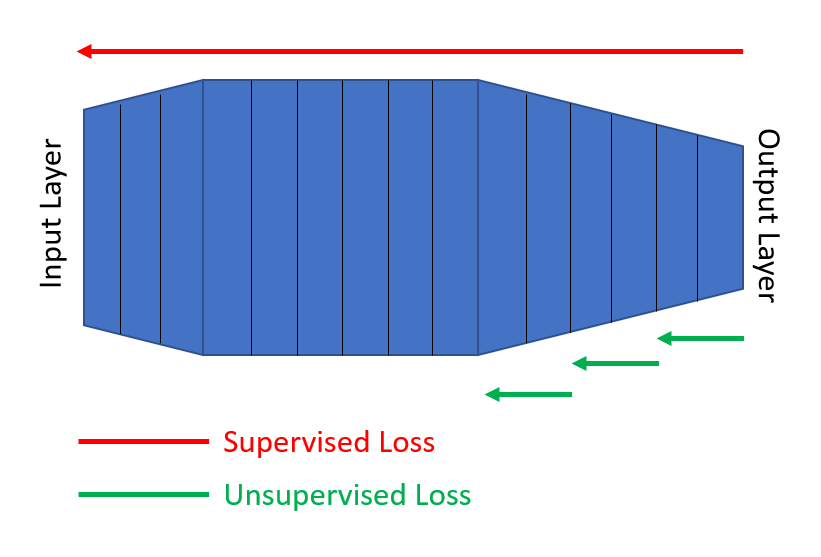}
   \vspace{-1cm}
   \caption{Proposed mechanism for combining supervised and unsupervised loss terms: the supervised loss affects all parameters in the network, whereas the unsupervised losses only affect weights in neighboring downstream layers.}
   \label{fig:losses}
\end{figure}

As stated, we aim to achieve memory-efficient learning of deep neural networks. In practice, we attain this goal by: (i) substituting wide fully connected (FC) layers with multiple, narrowing FC layers, and (ii) enforcing a regularization term in order to guide the learning process. As such, the learning objective comprises two elements, an unsupervised, geodesic-distance preserving loss~\cite{8953348} and a supervised, task-specific loss. For purposes of demonstration, we implement networks aimed at performing multi-class classification, so the supervised portion of the objective comprises the traditional cross-entropy loss. We believe that our framework is flexible enough to be able to operate in other supervised tasks including object detection and semantic segmentation. As illustrated in Fig.~\ref{fig:losses}, the supervised loss is backpropagated all the way through the network from the output layer; in contrast, the unsupervised component is broken down into local losses which only affect neighboring upstream layers in the bottleneck portion of the network. This design is motivated by the following observations: (i) the mapping between intermediate layers in the bottleneck section of the network is a progressive dimensionality reduction process, and distance preservation is only required between spaces that are adjacent to each other dimensionality-wise~\cite{8953348}; (ii) the feature space spanned by the activations in the deeper layers should best approximate the true data manifold~\cite{NIPS2017_0ebcc77d}; and, (iii) the concept of distance is not as well defined in feature spaces closer to the data space~\cite{pmlr-v28-bengio13}. 

We next formalize the proposed learning framework. Let $X={x_0,x_2,\dots,x_{M-1}}$ denote the variables representing the raw input data as it lies on the ambient space. We propose to learn a non-linear mapping $f(x;\theta)$ --effected by a deep neural network-- between the ambient space and the intrinsic space by finding optimal parameters $\theta^*$ according to the following objective function:

\begin{equation}
\theta^* = \argmin\limits_{\theta} \{ \mathcal{L}_s(\theta) + \mathcal{L}_u(\theta) \}
\label{eq:loss}
\end{equation}

\noindent where $\mathcal{L}_s(\theta)$ and $\mathcal{L}_u(\theta)$ are the supervised and unsupervised loss terms, respectively. When the supervised task at hand is classification, cross-entropy can be used in place of the supervised term in Eq.~\ref{eq:loss}, namely:

\begin{equation}
\mathcal{L}_s(\theta) = -\sum_{n=0}^{L-1}y_n\text{log}f(x_n;\theta)
\label{eq:losss}
\end{equation}

\noindent where $X={x_0,x_2,\dots,x_{L-1}}$ denotes the set of samples in the mini-batch (with $L<M$), $Y={y_0,y_1,\dots,y_{L-1}}$ the corresponding labels, and $f(x_n;\theta)$ the output of the network (parameterized by $\theta$) to input $x_n$.

Further, let $k=0, 1, ..., K-1$ denote the indices of the layers comprising the bottleneck section of the network. Then 
\vspace{-0.1cm}
\begin{equation}
\mathcal{L}_u(\theta) = \sum_{k=1}^{K-1}\alpha_k\mathcal{L}_u^{(k)}(\theta^{(k)})
\label{eq:lossu}
\end{equation}

\noindent where $\mathcal{L}_u^{(k)}(\theta)$ is the distance-preserving loss between the activations from the $k$-th and $(k-1)$-th layers, $\theta^{(k)}$ are the network parameters involved in the optimization (i.e., the parameters of layers $k$ and $k-1$), and $\alpha_k$ are the weights determining the contribution of each loss term. In our case,
\vspace{-0.2cm}
\begin{equation}
\begin{split}
\mathcal{L}_u^{(k)}(\theta^{(k)}) =\\
\sum_{n=0}^{L-1}\sum_{m=0}^{L-1}[d(x_n^{(k)},x_m^{(k)})-d(x_n^{(k-1)},x_m^{(k-1)})]^2+\lambda\parallel\theta^{(k)}\parallel^2 
\label{eq:lossui}
\end{split}
\end{equation}

\noindent where $d(\cdot)$ is the Euclidean distance operator, $L$ is the number of samples in the mini-batch, $x_n^{(k)}$ is the $k$-th layer activation corresponding to training sample $x_n$, and the last term is a regularizer on the parameters of the layers involved, controlled by weight $\lambda$. Note that $x_n^{(k)}$ is a function of $\theta^{(k)}$, but this dependence has been omitted in the notation for simplicity.

\section{Experimental Results}

\subsection{Datasets}
We evaluate the performance of the proposed framework on the CIFAR-10 and CIFAR-100 datasets. The CIFAR-10 dataset is a compilation of pictures used as a standard evaluation method for machine learning algorithms that tackle image classification tasks. The collection comprises $60\,000$, $32\times32$-pixel color images belonging to 10 categories corresponding to everyday-life objects. Each category has $6\,000$ images that are distinct. The dataset is partitioned into 50,000 training and 10,000 test images. The CIFAR-100 dataset is similar to CIFAR-10 except that it has $100$ classes with $600$ images each, similarly partitioned into $500$- and $100$-image training and test sets per class.  

\subsection{Experimental Setup}
\begin{figure*}[t]
  \centering
   \includegraphics[width=0.9\linewidth]{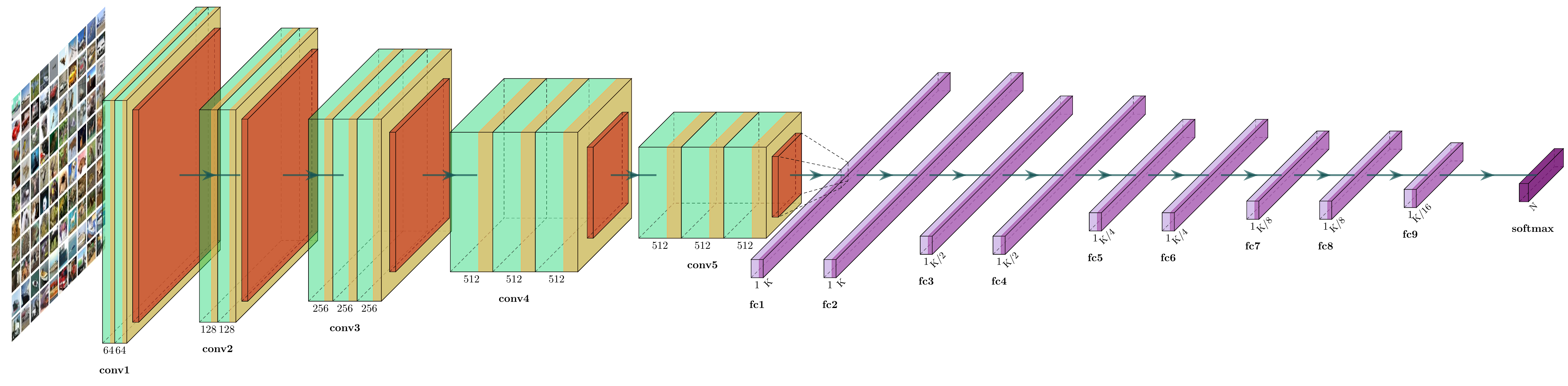}
   \caption{Leveraged network architecture with VGG16 as a feature extractor. The largest dimension of the fully connected layers is set to $K$ and decreased by a factor $1/2$ bilayerly.}
   \label{fig:vgg16}
\end{figure*}

We leverage the convolutional section of the VGG16 model~\cite{Simonyan15} as a feature extractor. The original VGG16 architecture consists of 16 convolutional and pooling layers followed by three fully connected layers. In order to demonstrate the efficacy of the proposed framework, we decrease the memory footprint of VGG16 by replacing the fully connected section of the network with a bottleneck structure. Specifically, we introduce a cascade of fully connected layers, each halving in width until a width commensurate with the intrinsic dimensionality of the dataset at hand is reached. This design choice is motivated by a desire to improve the inductive bias of the architecture. Figure~\ref{fig:vgg16} illustrates an example of such an architecture.

The manifold-oriented regularization term is implemented in the form of distance-preserving objectives across successive layers in the bottleneck portion of the network. As described earlier, the regularizer objective encourages the mapping between adjacent spaces to preserve pairwise distances between the activations. More specifically, let $x_n$ and $x_m$ be two data samples in a mini-batch and $x_n^{(k)}$ be the $k$-th layer activation corresponding to sample $x_n$. Then, as described by Eq.~\ref{eq:lossui}, distances between activations for $x_n$ and $x_m$ in the $k$-th layer are encouraged to match the distances between the activations for the same data points in the previous, or $(k-1)$-th layer. As stated, we use a combination of supervised and unsupervised losses with different scopes: while the supervised loss from Eq.~\ref{eq:loss} affects all the parameters in the network, the unsupervised objectives from Eq.~\ref{eq:lossu} is backpropagated every two layers. This design choice is aimed at enabling a progressive dimensionality reduction process through mappings that preserve the manifold structure of the data across the different intermediate spaces. The regularization weights for the regularization objective enforced across the first half of the bottleneck region is set at a smaller value than those that influence the second half. This design choice is motivated by the fact that the feature space spanned by the activations closer to the final layer should best approximate the true data manifold~\cite{NIPS2017_0ebcc77d}, and the concept of a semantically meaningful distance is less well defined with features closer to the data space~\cite{pmlr-v28-bengio13}. Lastly, the number of epochs and learning rate are set to $200$ and $0.01$, respectively.

\subsection{Quantitative Performance Evaluation}
We measure the effectiveness of the proposed framework in terms of absolute accuracy (larger is better) and in the form of empirical generalization error, which is defined as the difference between the performance on the training set and the performance on the test set (smaller is better). To that end, we train and test classifiers with and without the layer-wise manifold regularization from Eqs.~\ref{eq:lossu} and \ref{eq:lossui} on both the CIFAR-10 and CIFAR-100 datasets. In order to gauge the effect of different bottleneck widths and mini-batch sizes, we perform runs with different combinations of said parameters on CIFAR-10. We adjust the width of the narrowest layer of the bottleneck section of the network, and denote it as $W$. In the CIFAR-10 experiments, we perform tests with $W=\{8, 16, 32\}$, as well as mini-batch sizes of $5$, $7$ and $9$. Tables \ref{tab:cifar10} and \ref{tab:delta} contain the results on absolute performance and empirical generalization error, respectively. 


Experimental results indicate that when a manifold-oriented regularizer is employed, the resulting networks showcase improved resilience to architectural changes (in this case in particular, the narrowest width of the bottleneck section of the network, $W$), as well as hyperparameters such as mini-batch size. In terms of absolute accuracy, as recorded in Table \ref{tab:cifar10}, the performance of the regularized network is better than that of its non-regularized version across the board. In particular, the performance of the network that doesn't leverage regularization seems to suffer as $W$ and mini-batch size decrease. We observe a similar effect in the empirical generalization error numbers from Table \ref{tab:delta}, which grow to almost $50\%$ in the most extreme case, namely, the one with the narrowest bottleneck and the smallest mini-batch. These numbers point to extreme overfitting taking place, likely due to the fact that a sound data manifold representation is extremely difficult to find in such a restrictive learning environment. Impressively, the proposed framework maintains the empirical generalization error under $10\%$ across the board. In order to determine a baseline, we ran similar experiments with the vanilla VGG16 network, as shown in Table \ref{tab:origvggCIFAR10}. As the mini-batch size decreases from the standard of $64$ and into single-digit territory, performance suffers significantly, in particular in the extremely small mini-batch size regime.

\begin{table}[t]
\centering
\begin{tabular}[t]{ l c c c}
& \multicolumn{3}{c}{Bottleneck Width ($W$)} \\
\hline
\multicolumn{1}{l}{Method$\downarrow$}&\multicolumn{1}{c}{$32$} & \multicolumn{1}{c}{$16$} &  \multicolumn{1}{c}{$8$} \\
\hline
\multicolumn{4}{c}{Mini-batch Size=$9$}\\
\hline
Without regularizer&$83.67$&$83.47$&$70.04$\\
With regularizer&$\mathbf{85.70}$&$\mathbf{86.27}$&$\mathbf{86.03}$\\
\hline
 \multicolumn{4}{c}{Mini-batch Size=$7$}\\
\hline
Without regularizer&$62.25$&$81.40$&$82.70$\\
With regularizer&$\mathbf{75.43}$&$\mathbf{87.83}$&$\mathbf{86.55}$\\
\hline
\multicolumn{4}{c}{Mini-batch Size=$5$}\\
\hline
Without regularizer&$50.30$&$56.25$&$26.10$\\
With regularizer&$\mathbf{73.80}$&$\mathbf{67.71}$&$\mathbf{75.60}$\\
\hline
\end{tabular}
\caption{Classification performance (in $\%$, higher is better) of competing frameworks on CIFAR-10 for different values of mini-batch size and $W$.}
\label{tab:cifar10}
\end{table}
\begin{table}[t]
\centering
\begin{tabular}[t]{ l c c c}
& \multicolumn{3}{c}{Bottleneck Width ($W$)} \\
\hline
\multicolumn{1}{l}{Method$\downarrow$}&\multicolumn{1}{c}{$32$} & \multicolumn{1}{c}{$16$} &  \multicolumn{1}{c}{$8$} \\
\hline
\multicolumn{4}{c}{Mini-batch Size=$9$}\\
\hline

Without regularizer&$9.67$&$6.66$&$18.41$\\
With regularizer&$\mathbf{4.06}$&$\mathbf{4.26}$&$\mathbf{7.56}$\\
\hline
 \multicolumn{4}{c}{Mini-batch Size=$7$}\\
\hline

Without regularizer&$24.65$&$16.53$&$9.20$\\
With regularizer&$\mathbf{9.39}$&$\mathbf{4.49}$&$\mathbf{4.01}$\\
\hline
\multicolumn{4}{c}{Mini-batch Size=$5$}\\
\hline
Without regularizer&$17.82$&$19.00$&$48.53$\\
With regularizer&$\mathbf{8.00}$&$\mathbf{7.01}$&$\mathbf{5.70}$\\
\hline
\end{tabular}
\caption{Empirical generalization error (in $\%$, lower is better) of competing frameworks on CIFAR-$10$ for different values of mini-batch size and $W$.}
\label{tab:delta}
\end{table}

\begin{table}
  \centering
  \begin{tabular}{@{}lccccc@{}}
  & \multicolumn{5}{c}{Mini-batch Size} \\
\hline
    Method$\downarrow$&64 & 16&9 & 7 & 5\\
      \hline
  \multicolumn{6}{c}{Classification Performance}\\
    \hline
    VGG16&$92.86$&$75.84$& $70.74$& $55.27$ & $33.47$\\
  \hline
      \hline
  \multicolumn{6}{c}{Generalization Error}\\
\hline  
    VGG16&$6.01$&$4.01$& $4.30$& $6.76$ & $10.99$\\
    \bottomrule
\end{tabular}
  \caption{Baseline classification performance ($\%$) and generalization error ($\%$) of the vanilla VGG16 network on CIFAR-10 as a function of mini-batch size.}
  \label{tab:origvggCIFAR10}
\end{table}

We performed similar experiments on CIFAR-100 with a fixed mini-batch size of $5$ and varying values of $W$, as the results from Table \ref{tab:bestcifar100} indicate. As before, the network trained with manifold-based regularization outperforms the unregularized network consistently. Absolute performance is better across the board, while empirical generalization performance remains competitive. While for $W=32$ the unregularized network showcases better generalization capabilities, this is at the cost of pretty abysmal absolute performance to begin with. Compare the figures from Table \ref{tab:bestcifar100} with the baseline accuracy of $17.87\%$ of the vanilla VGG16 network when trained with a mini-batch size of 5, which is inferior to all of the performance numbers obtained with our proposed framework, regardless of the choice of $W$. For the same network, under the same training conditions, the generalization error is of $9.68\%$, which, while in line with the generalization error of the proposed framework, is proportionally much larger given the absolute performance of the model on the test set. While the memory savings of the proposed model won't be quantified until the next section of the paper, we highlight at this point that the networks produced with the proposed framework consume less than half the GPU memory than the traditional, vanilla version of the baseline network, in this case VGG16.

\begin{table}
  \centering
  \begin{tabular}{@{}lccc@{}}
  & \multicolumn{3}{c}{Bottleneck Width ($W$)} \\
 \hline
\multicolumn{1}{l}{Method$\downarrow$}&\multicolumn{1}{c}{$32$} & \multicolumn{1}{c}{$16$} &  \multicolumn{1}{c}{$8$} \\
  \hline
  \multicolumn{4}{c}{Classification Performance}\\
\hline
    Without regularizer&$14.84$& $29.74$& $15.40$ \\
    With regularizer&$\mathbf{37.00}$& $\mathbf{33.99}$& $\mathbf{18.74}$ \\
    \hline
    \hline
    \multicolumn{4}{c}{Generalization Error}\\
\hline
    Without regularizer&$\mathbf{5.10}$& $13.82$& $13.41$ \\
    With regularizer&$8.36$& $\mathbf{12.55}$& $\mathbf{12.87}$ \\
    \hline
  \end{tabular}
  \caption{{Classification performance (in $\%$, higher is better) and empirical generalization error (in $\%$, lower is better) of competing frameworks on CIFAR-100 with a mini-batch size of $5$ and different values of $W$.}}
  \label{tab:bestcifar100}
\end{table}

\subsection{Memory Profiling Results}
In this section, we compare the memory consumption patterns between vanilla networks and our proposed framework with a range of bottleneck widths and mini-batch sizes. We break down memory consumption into two categories, namely model- (including model parameter storage and resident buffer memory) and computation-related (including storage of data, activations and gradients, as well as ephemeral tensors and variables), in line with the taxonomy introduced in \cite{10.1145/3368089.3417050} (see Table 2 in the reference), albeit not as granular.

\begin{figure}[t]
    \centering
    \begin{minipage}[c]{0.49\linewidth}
    \includegraphics[width=\linewidth]{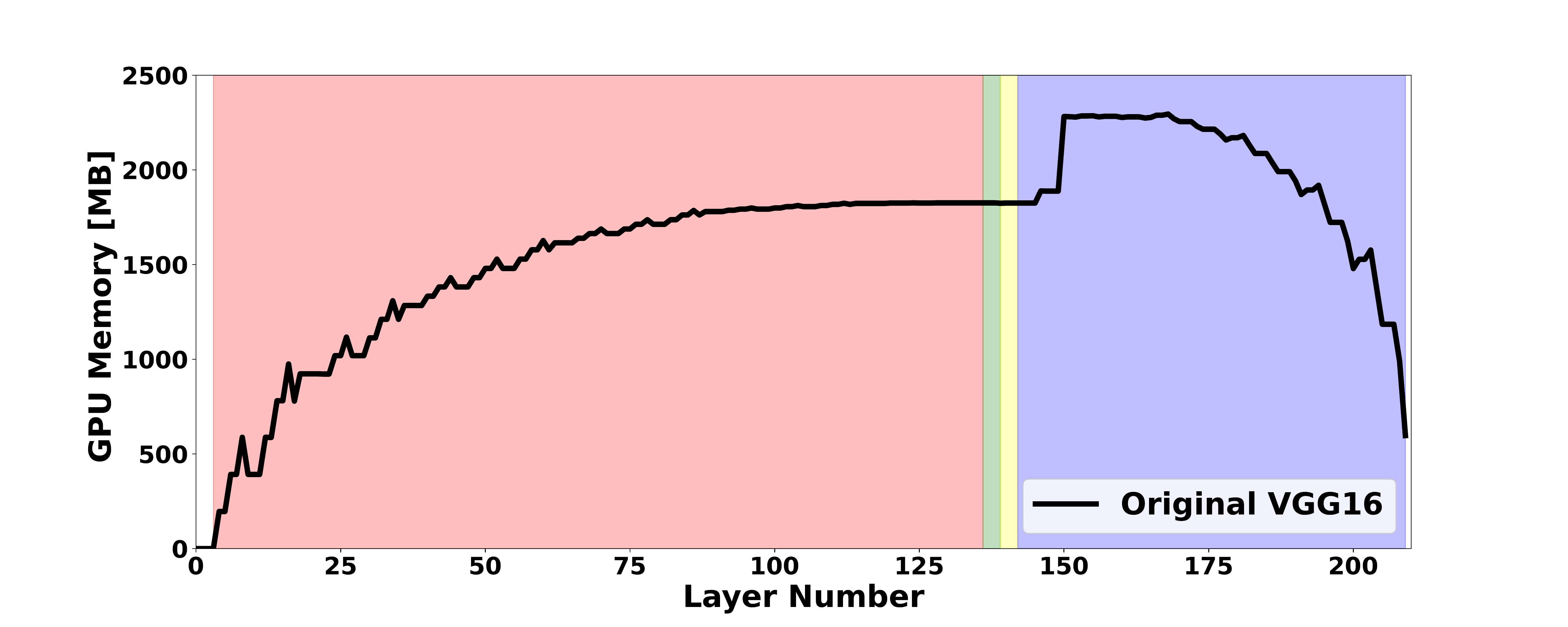}\\
    \centering{(a)}
    \end{minipage}
    \begin{minipage}[c]{0.49\linewidth}
    \includegraphics[width=\linewidth]{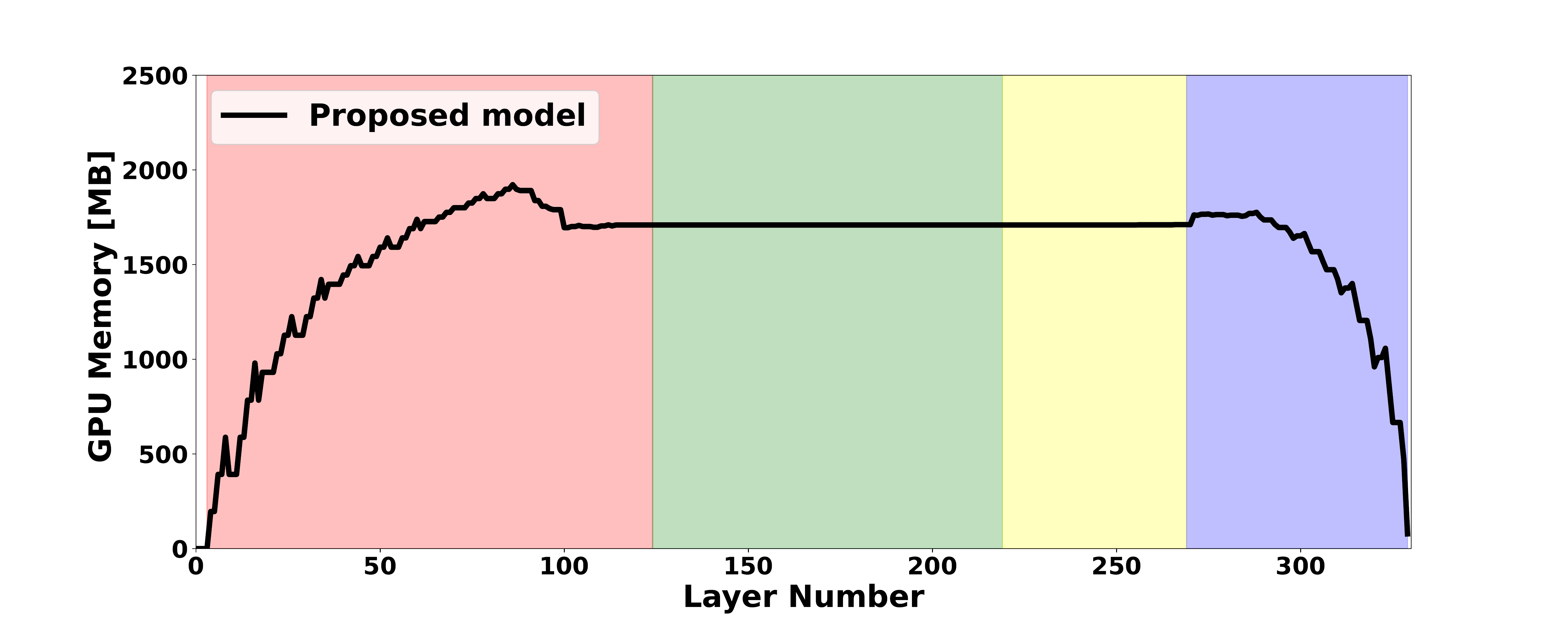}\\
    \centering{(b)}
    \end{minipage}
    \begin{minipage}[c]{0.49\linewidth}
    \includegraphics[width=\linewidth]{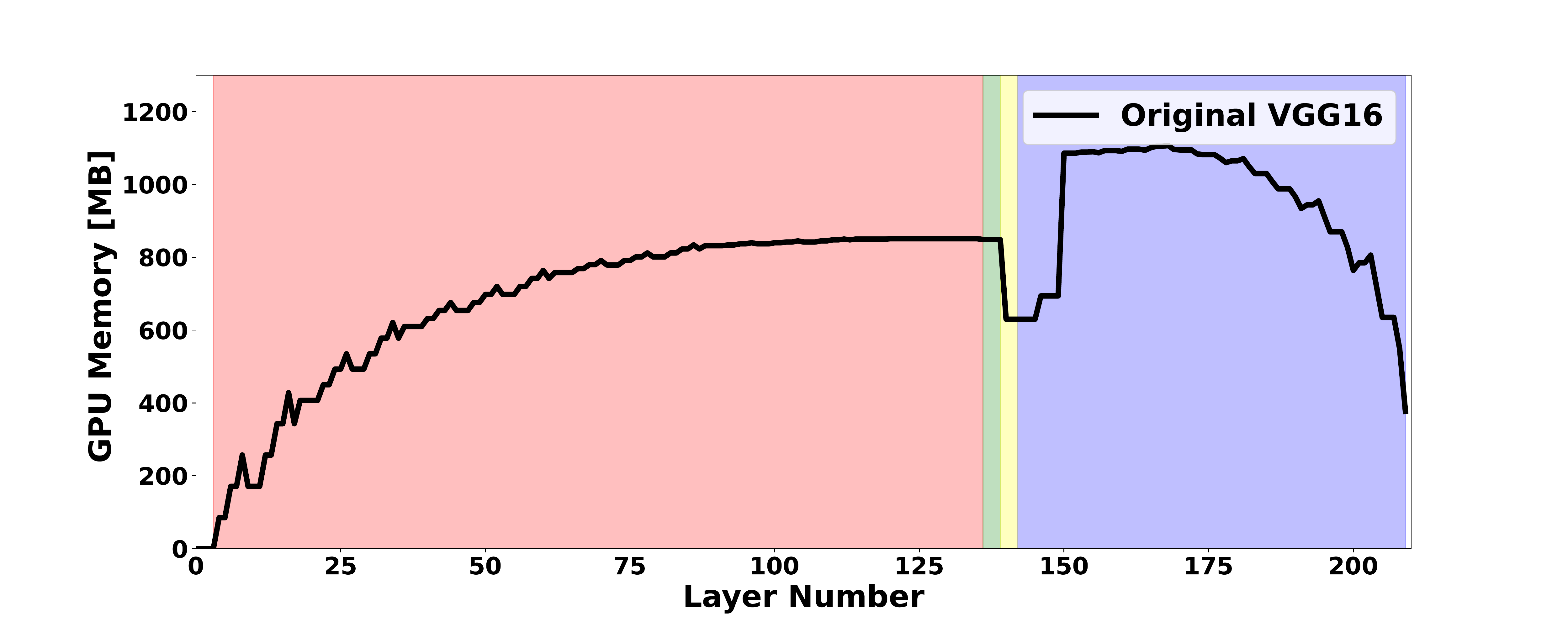}\\
    \centering{(c)}
    \end{minipage}
    \begin{minipage}[c]{0.49\linewidth}
    \includegraphics[width=\linewidth]{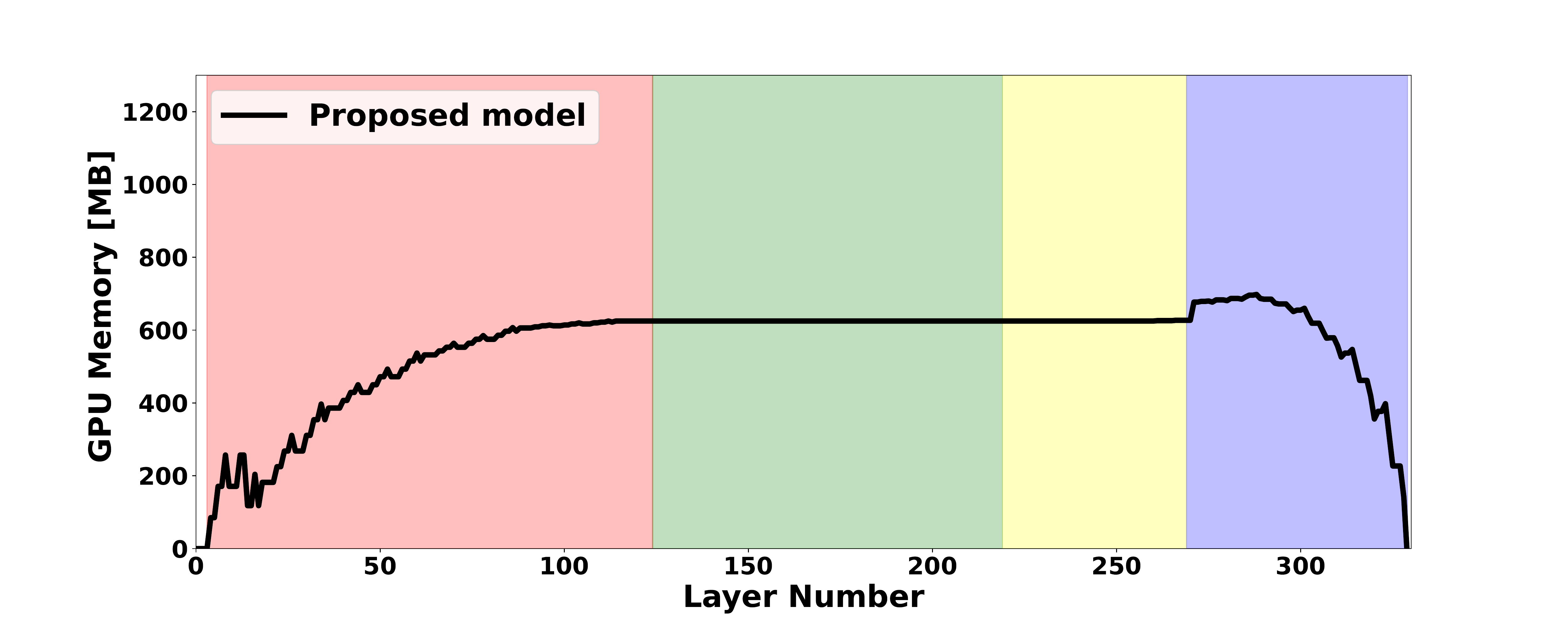}\\
    \centering{(d)}
    \end{minipage}
    \begin{minipage}[c]{0.49\linewidth}
    \includegraphics[width=\linewidth]{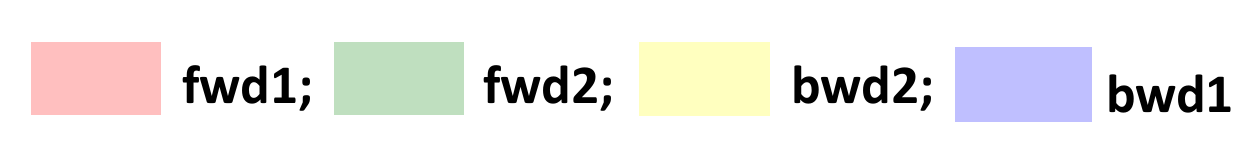}\\
    \end{minipage}
    \vspace{-10 pt}
    \caption{GPU memory usage as a function of time across a single training epoch: a) vanilla VGG16 with mini-batch size $16$, b) proposed method with mini-batch size $16$, c) vanilla VGG16 with mini-batch size $7$, and d) proposed method with mini-batch size $7$ and input size $3\times{}224\times{}224$. Stages \textbf{fwd1/bwd1} (in \textbf{red/purple}) and \textbf{fwd2/bwd2} (in \textbf{green/yellow}) correspond to memory consumption during the forward/backward passes through the convolutional and fully connected sections, respectively.}
    \label{fig:memorygpu}
\end{figure}

Table \ref{tab:memory} includes results relevant to model-related memory consumption. As expected, the fewer number of parameters enabled by the bottleneck section in the network lead to significant memory savings, in the order of $50\%$, almost independently of the bottleneck width. Figure~\ref{fig:memorygpu} shows plots illustrating dynamic memory consumption as networks are trained. In general, the pattern of behavior is as expected, namely, a slow increase in memory consumption is observed as the forward pass takes place until it stabilizes. Consumption decreases at a slighter faster rate as the backward pass is completed. In the vanilla VGG16 case in particular, we observe a sharp increase in the memory consumption shortly after the backward pass through the convolutional section starts. Two aspects of the plots are of particular importance: firstly, the dynamic memory consumption throughout is lower with the proposed framework (note that these savings are additional to the model-related memory savings from Table \ref{tab:memory}); secondly, peak memory consumption is markedly lower with the proposed method. This has significant implications to hardware requirements as peak memory consumption determines the size of the VRAM in the GPU required to train a model. Peak dynamic memory values for the vanilla VGG16 training process are 2,280 and 1,200 MBytes for batch sizes of 16 and 7, respectively; for the proposed framework, peak dynamic memory consumption stands at 1,750 and 800 MBytes for batch sizes of 16 and 7, respectively; these peaks take place during backpropagation through the fully connected portion of the network. For peaks that happen during the forward pass, vanilla VGG16 consumes 1,820 and 738 MBytes, while our proposed framework takes up 1,600 and 622 MBytes, with batch sizes of 16 and 7 respectively. Combining model and training memory consumption, the total observed peak is over $50\%$ smaller with our proposed framework relative to that reached by training the vanilla VGG16 network, even with extremely small mini-batch sizes. The memory savings will increase with an increasing mini-batch size. Lastly, note that traditional, \textit{post-hoc} network compression techniques can be applied to the resulting network for further memory savings at deployment.

\begin{table*}
  \centering
  \begin{tabular}{@{}l cccc@{}}
    {} & {} & \multicolumn{3}{c}{Proposed Framework -- Bottleneck Width ($W$)} \\
    \toprule
    {} & Vanilla VGG16 & $32$ & $16$ & $8$ \\
    \midrule
    Model Parameters & $134,586,664$&$62,059,304$&$60,480,040$&$60,082,856$\\
    Resident Buffer & $33,896$&$49,848$&$43,960$&$41,016$\\
    Total & $134,620,560$ &$62,109,152$&$60,524,000$ &$60,123,872$\\
    \bottomrule
  \end{tabular}
  \caption{Model-related memory consumption across different models (in bytes).}
  \label{tab:memory}
\end{table*}

\begin{figure*}[t]
    \centering
        \begin{minipage}[c]{0.15\linewidth}
    \centering{Conv Section}\\
    \includegraphics[width=\linewidth]{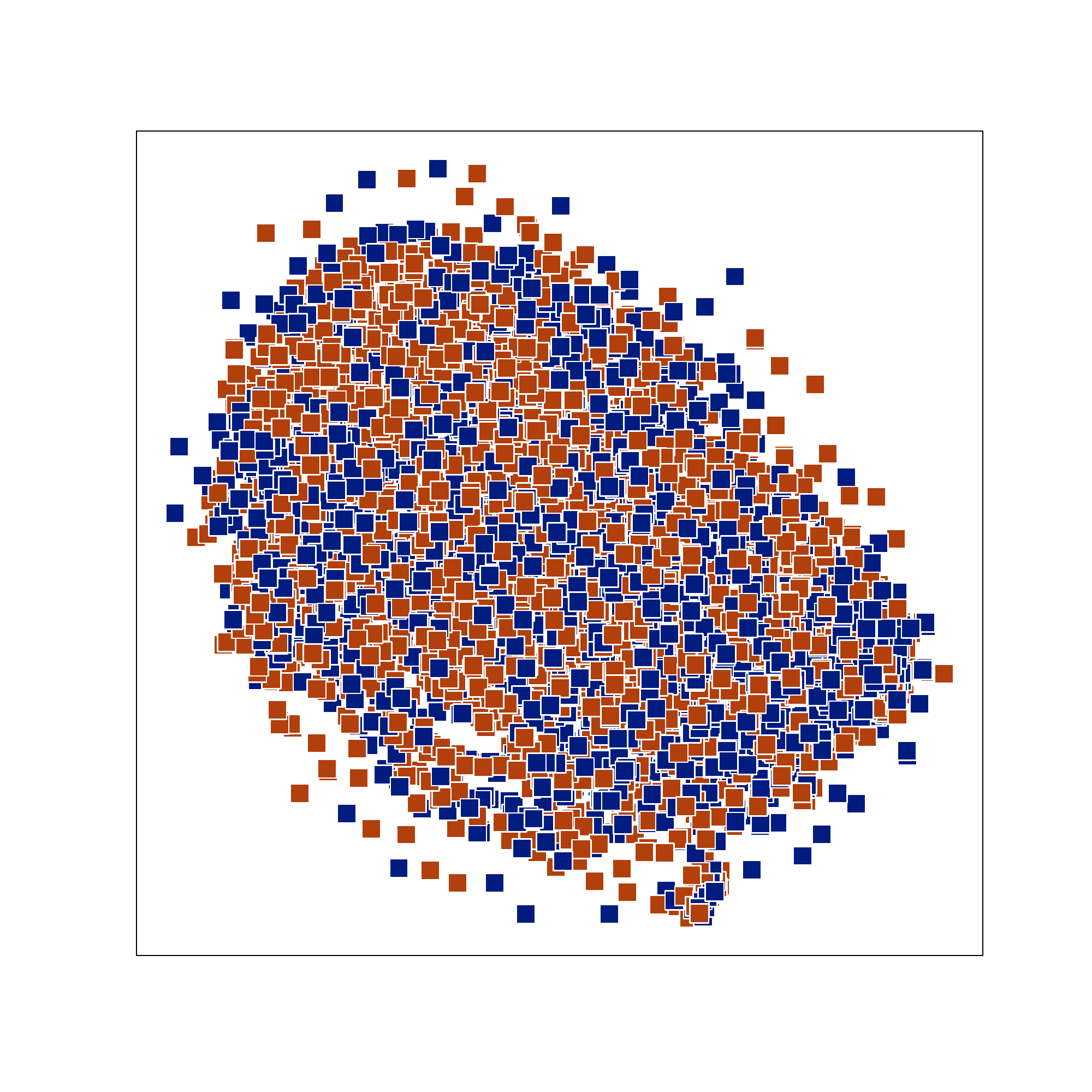}
    \end{minipage}
    \begin{minipage}[c]{0.15\linewidth}
    \centering{FC2}\\
    \includegraphics[width=\linewidth]{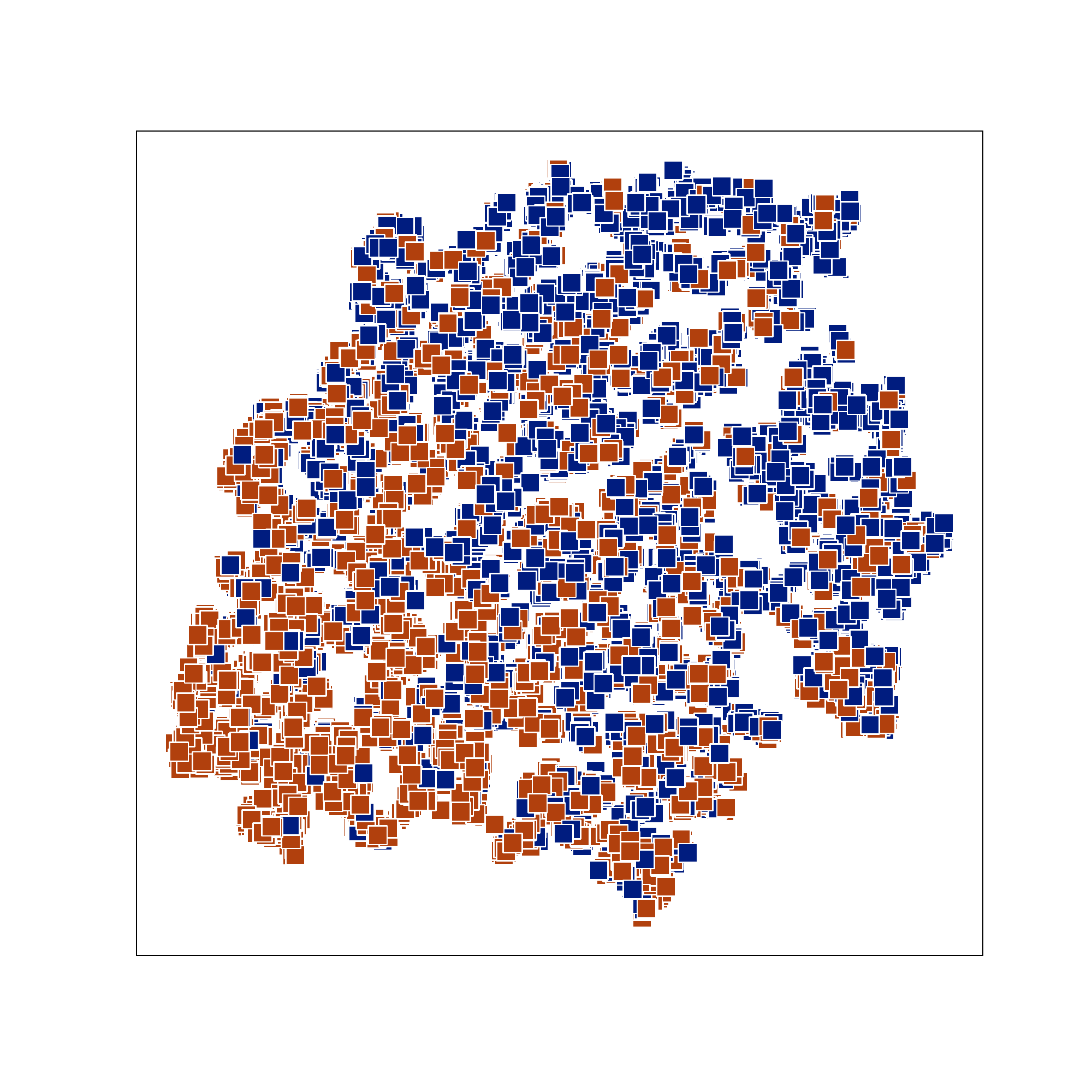}
    \end{minipage}
    \begin{minipage}[c]{0.15\linewidth}
    \centering{FC4}\\
    \includegraphics[width=\linewidth]{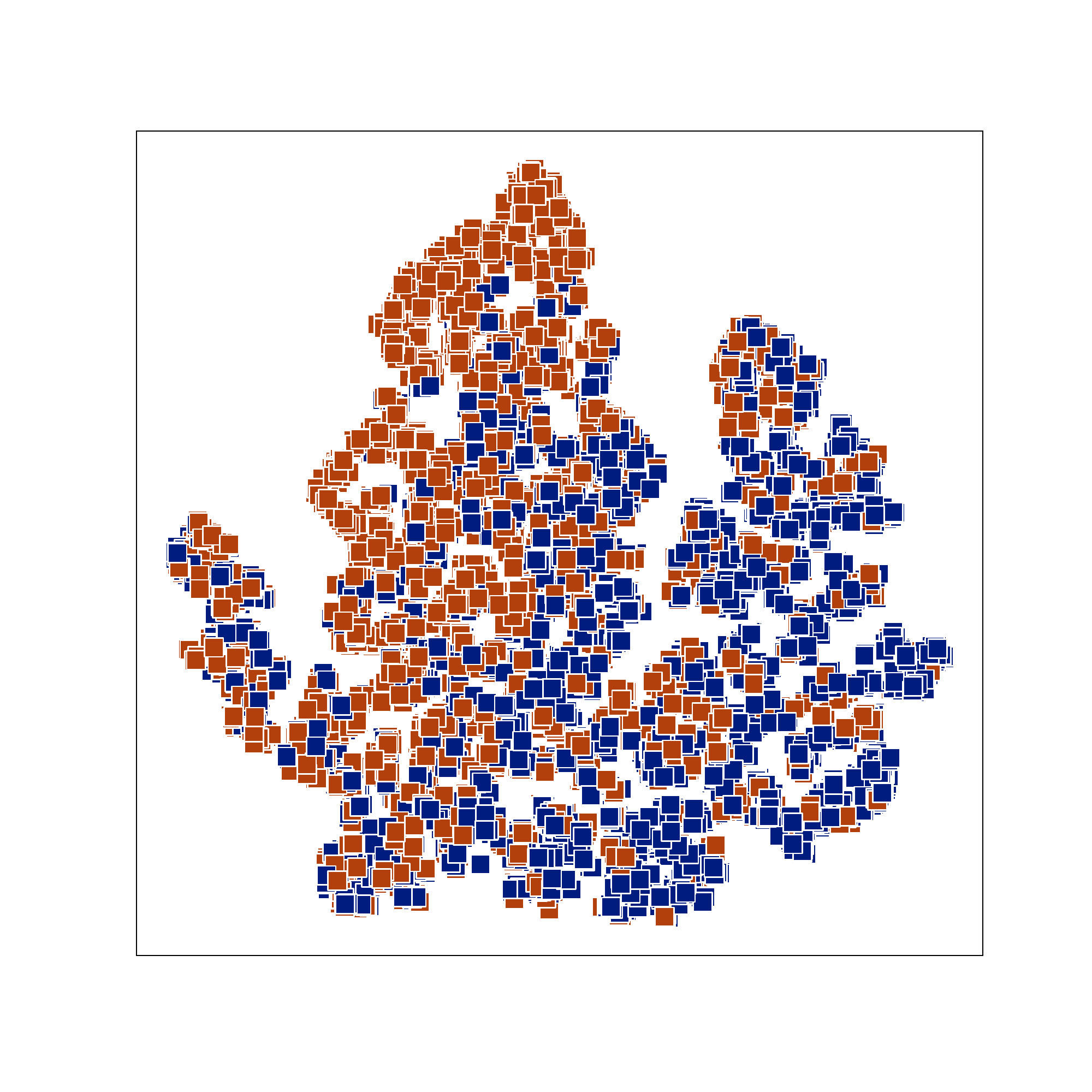}
    \end{minipage}
    \begin{minipage}[c]{0.15\linewidth}
    \centering{FC6}\\
    \includegraphics[width=\linewidth]{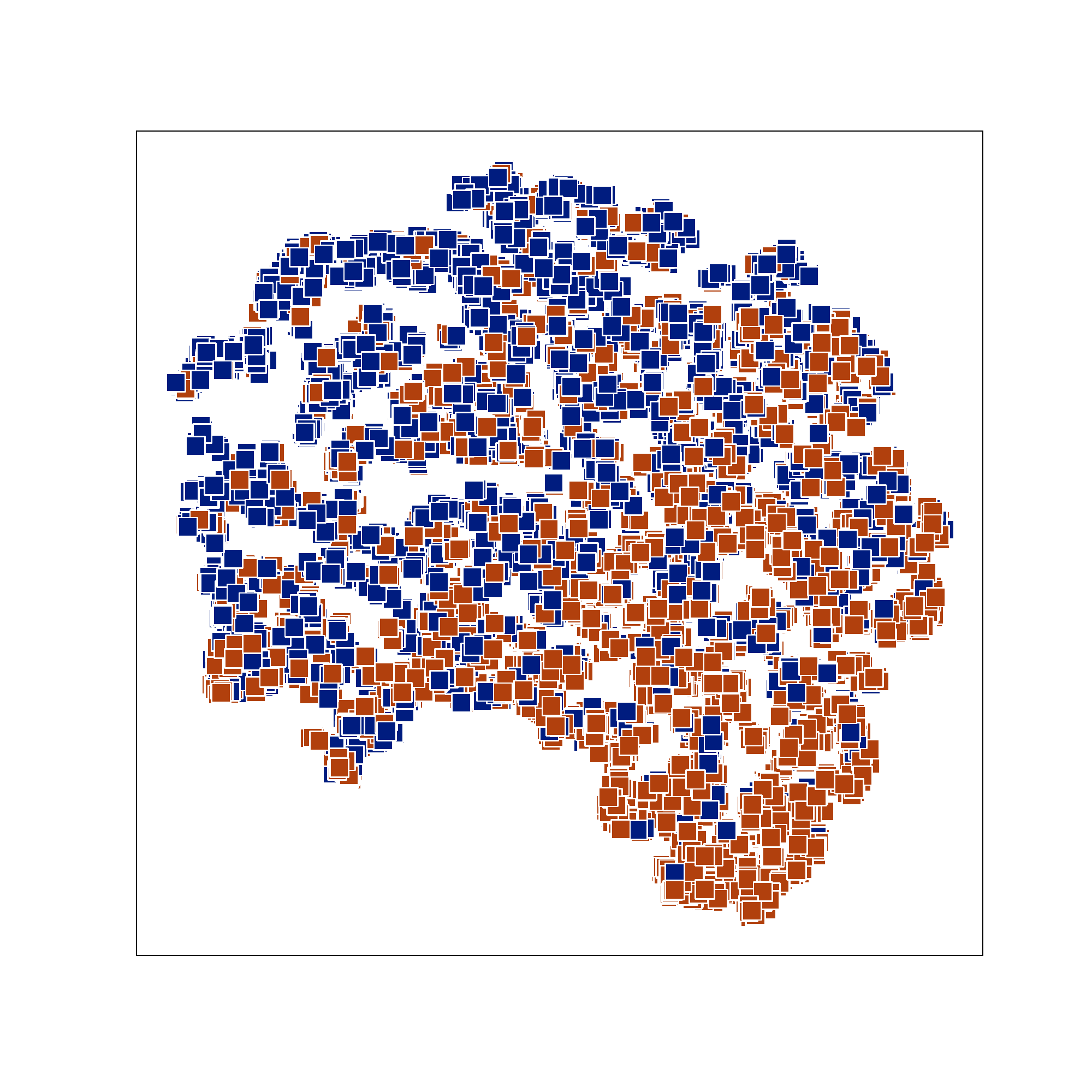}
    \end{minipage}
    \begin{minipage}[c]{0.15\linewidth}
    \centering{FC8}\\
    \includegraphics[width=\linewidth]{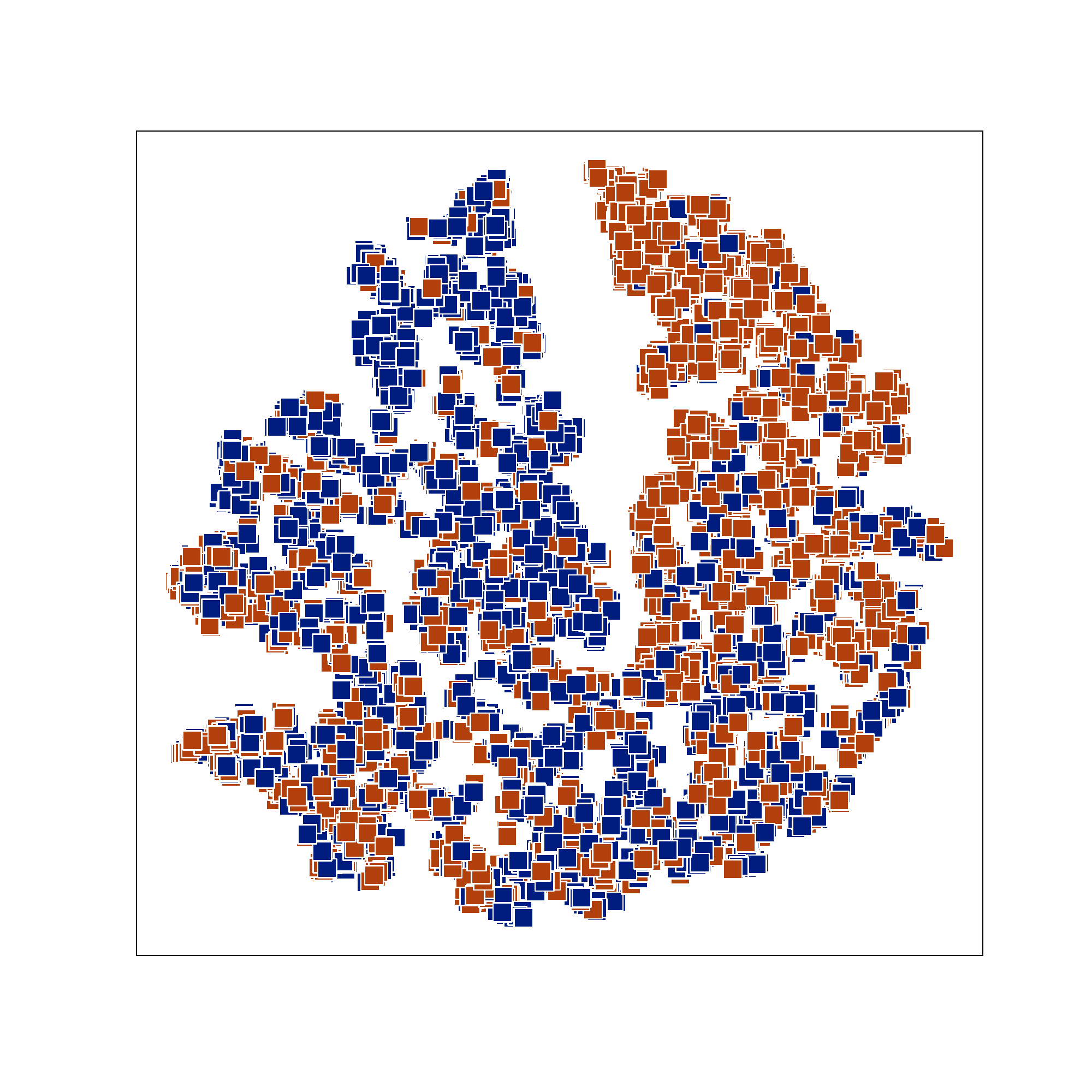}
    \end{minipage}
\begin{minipage}[c]{0.15\linewidth}
\centering{FC10}\\
    \includegraphics[width=\linewidth]{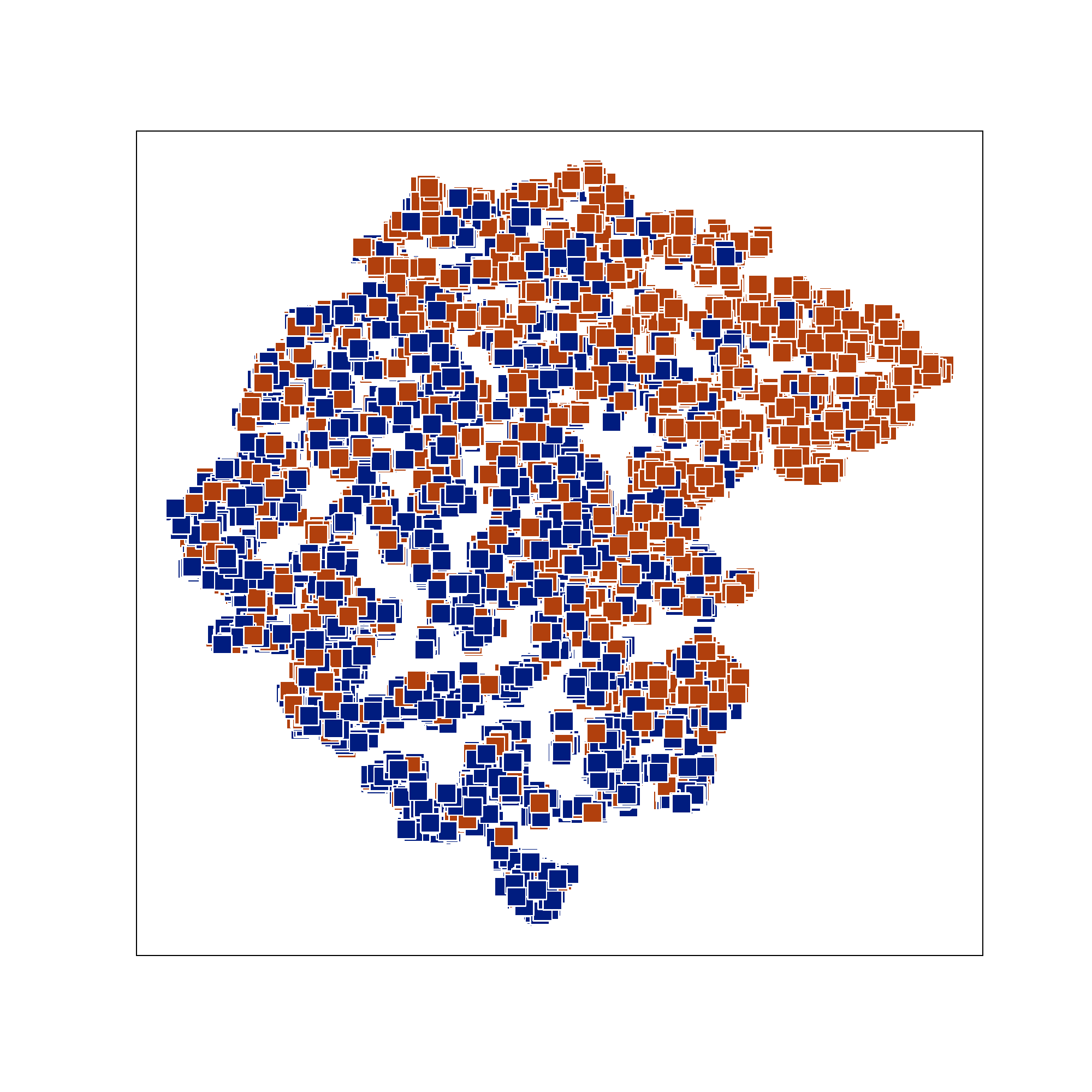}
    \end{minipage}
    \begin{minipage}[c]{\linewidth}
    \centering{a) Proposed approach without manifold regularization on training data.}\\
    \end{minipage}

      \begin{minipage}[c]{0.15\linewidth}
    \centering{Conv Section}\\
    \includegraphics[width=\linewidth]{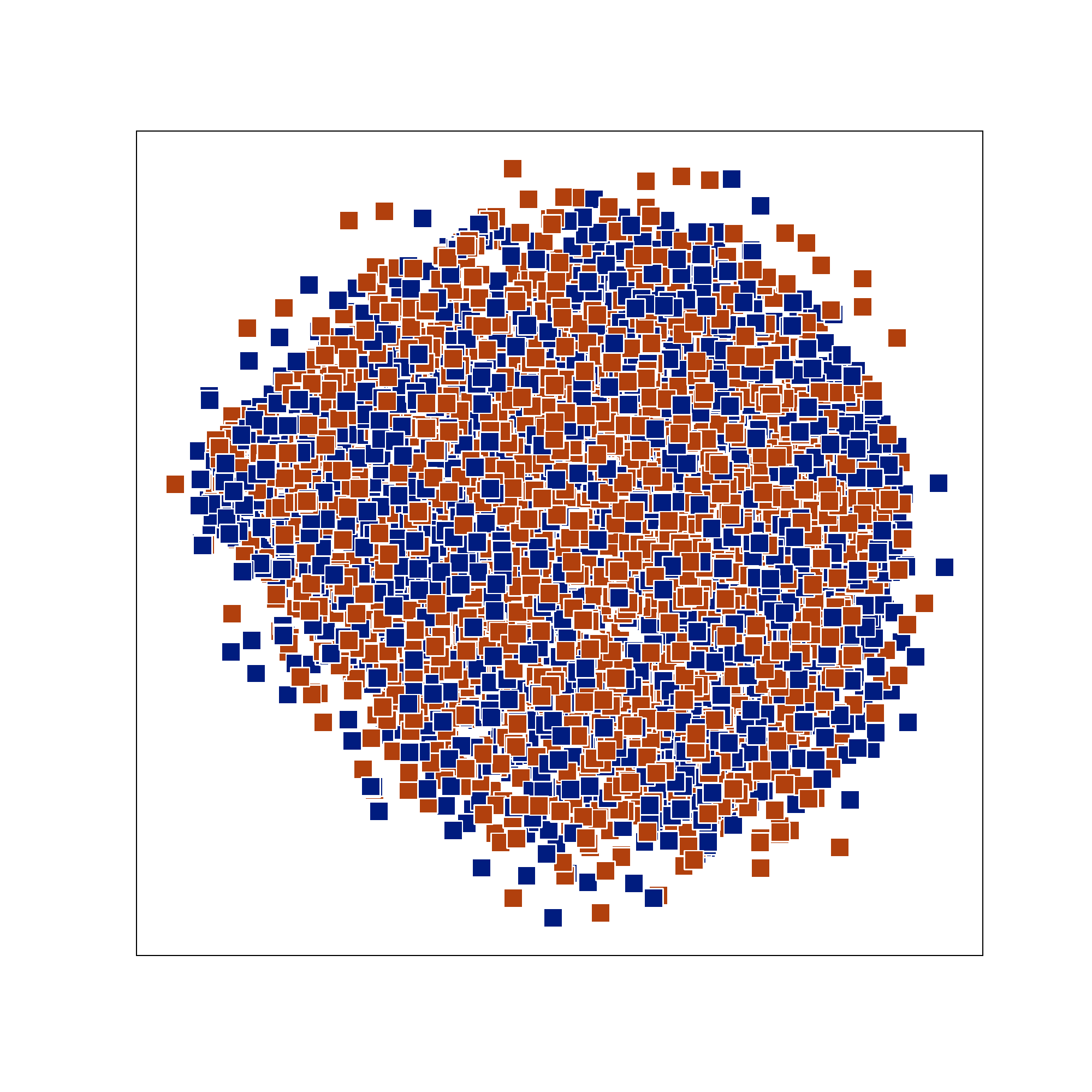}
    \end{minipage}
    \begin{minipage}[c]{0.15\linewidth}
    \centering{FC2}\\
    \includegraphics[width=\linewidth]{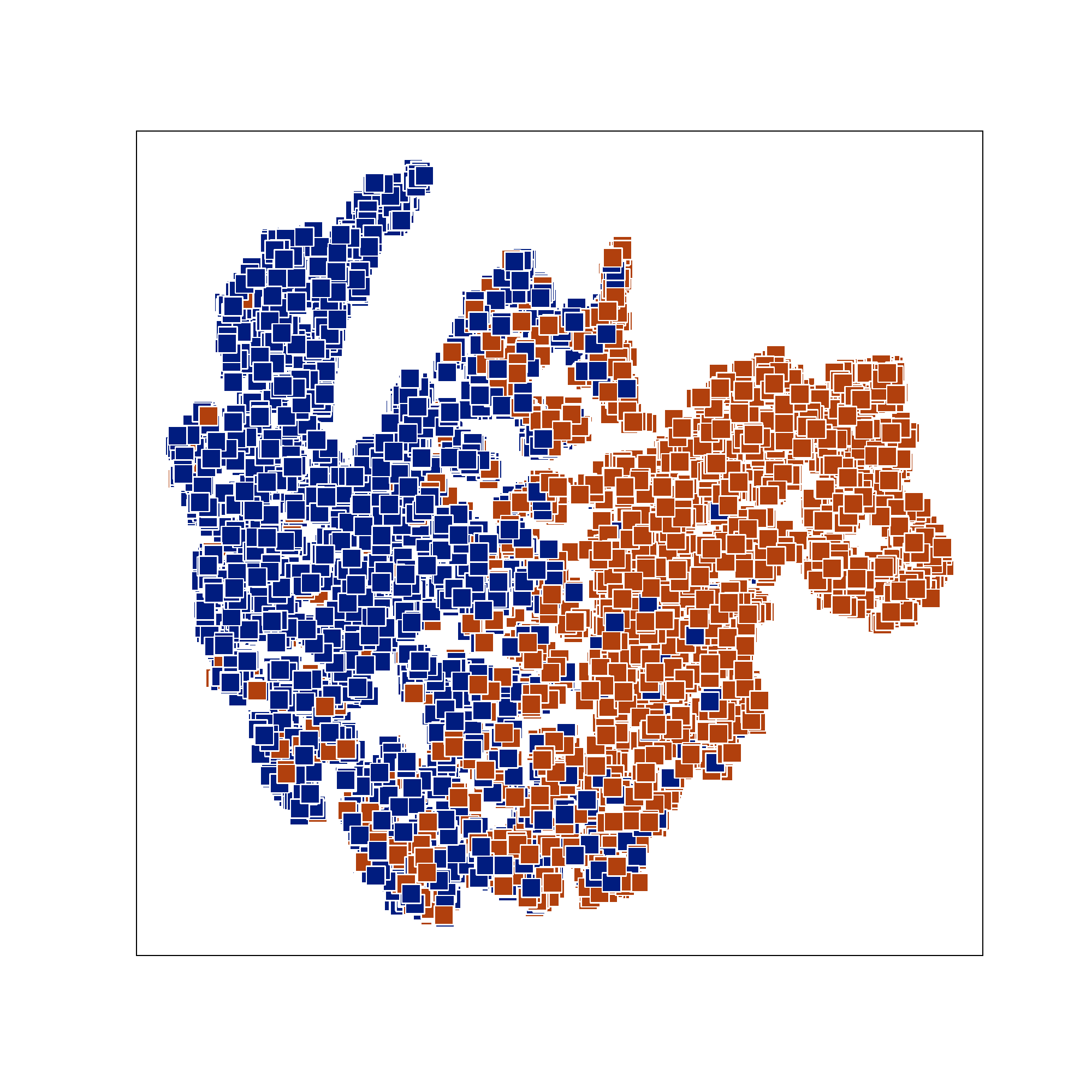}
    \end{minipage}
    \begin{minipage}[c]{0.15\linewidth}
    \centering{FC4}\\
    \includegraphics[width=\linewidth]{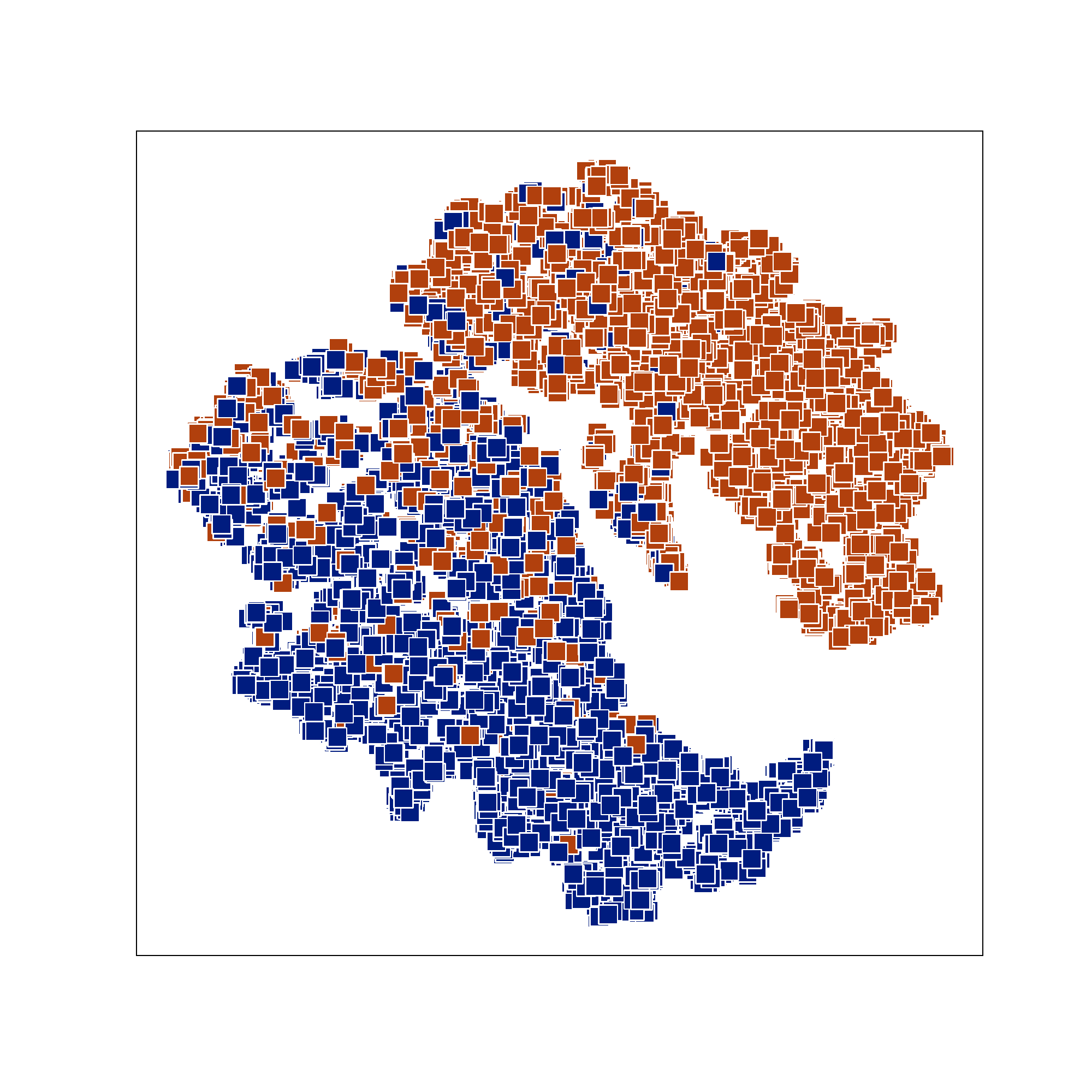}
    \end{minipage}
    \begin{minipage}[c]{0.15\linewidth}
    \centering{FC6}\\
    \includegraphics[width=\linewidth]{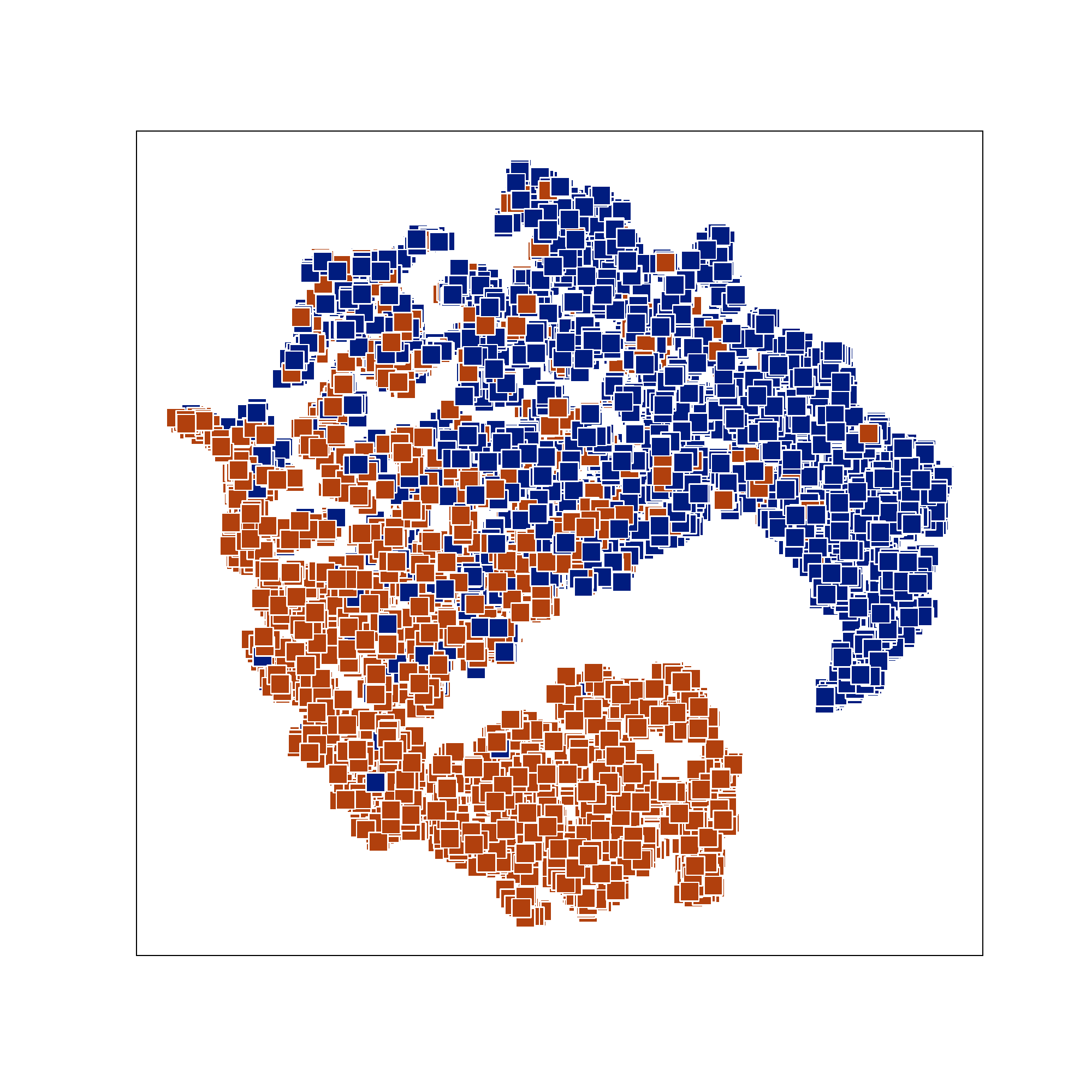}
    \end{minipage}
    \begin{minipage}[c]{0.15\linewidth}
    \centering{FC8}\\
    \includegraphics[width=\linewidth]{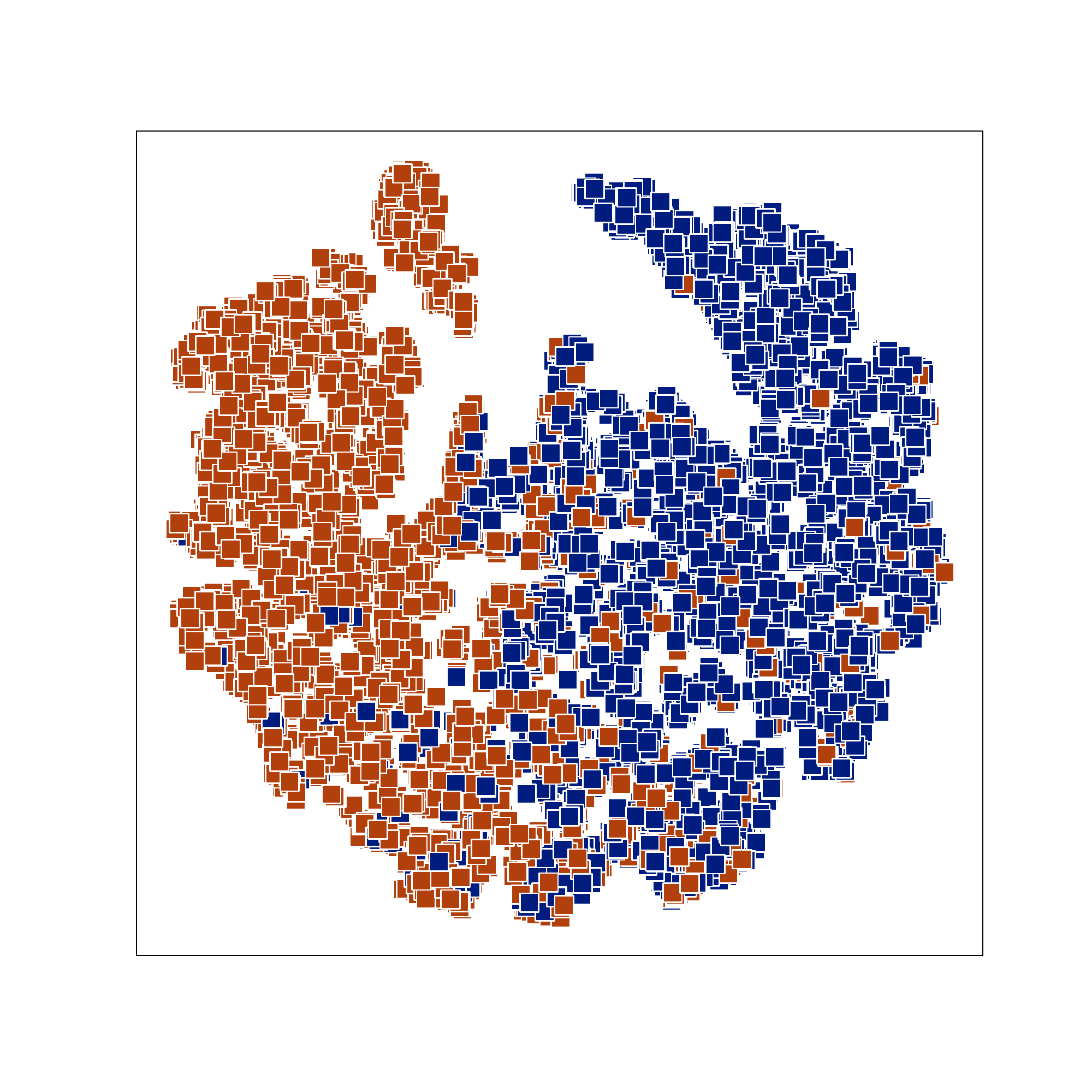}
    \end{minipage}
\begin{minipage}[c]{0.15\linewidth}
\centering{FC10}\\
    \includegraphics[width=\linewidth]{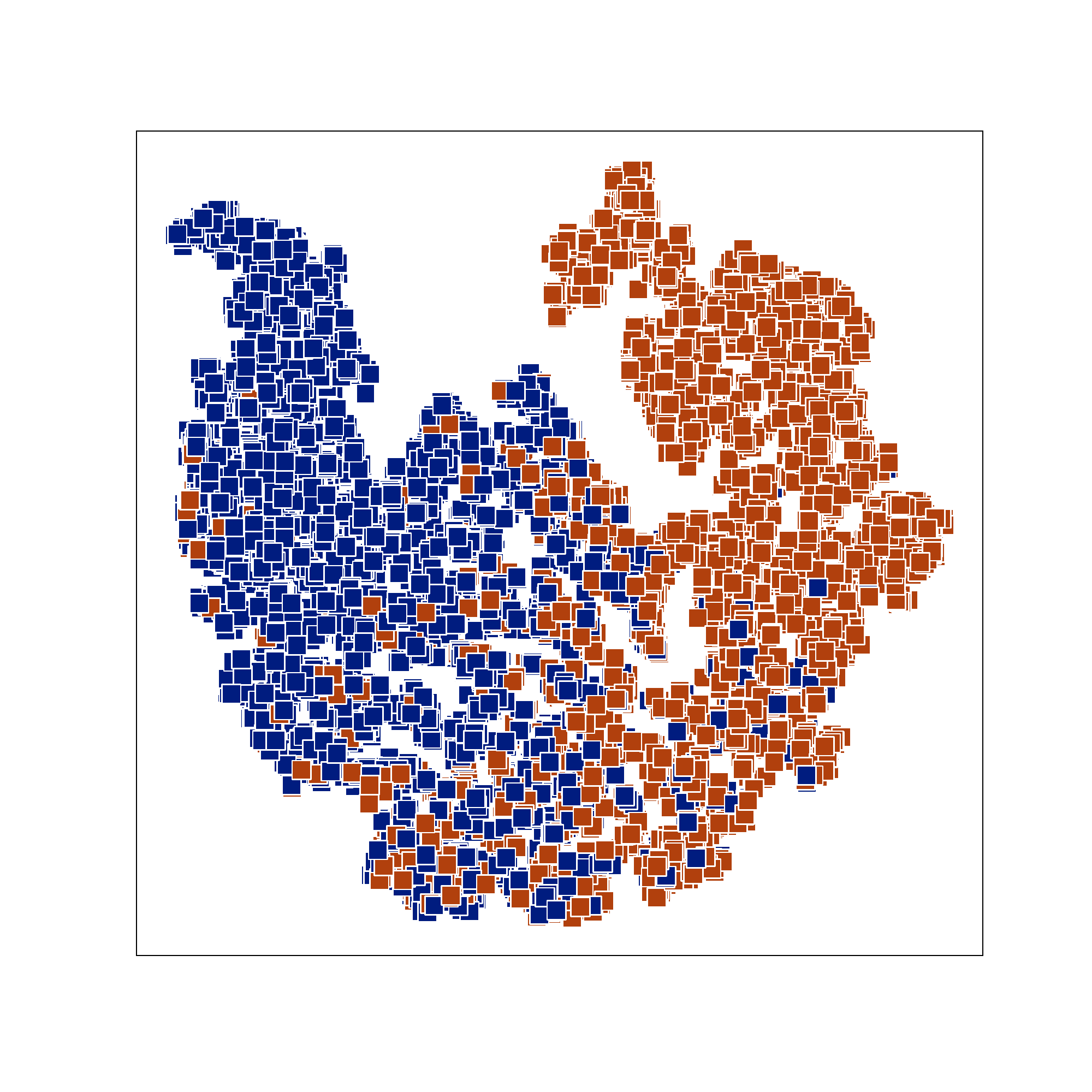}
    \end{minipage}
    \begin{minipage}[c]{\linewidth}
    \centering{b) Proposed approach with manifold regularization on training data.}\\
    \end{minipage}
     \begin{minipage}[c]{0.15\linewidth}
    \centering{Conv Section}\\
    \includegraphics[width=\linewidth]{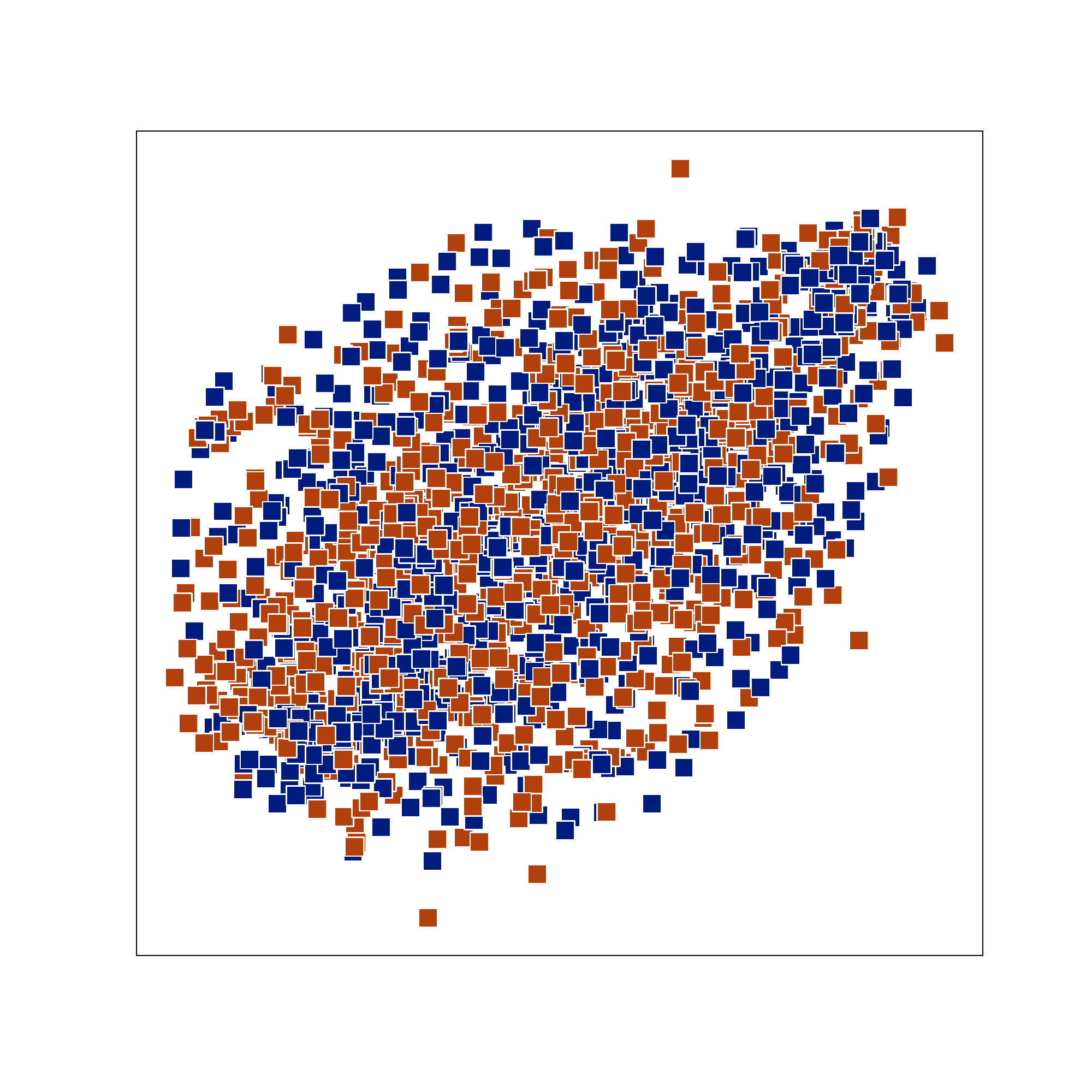}
    \end{minipage}
    \begin{minipage}[c]{0.15\linewidth}
    \centering{FC2}\\
    \includegraphics[width=\linewidth]{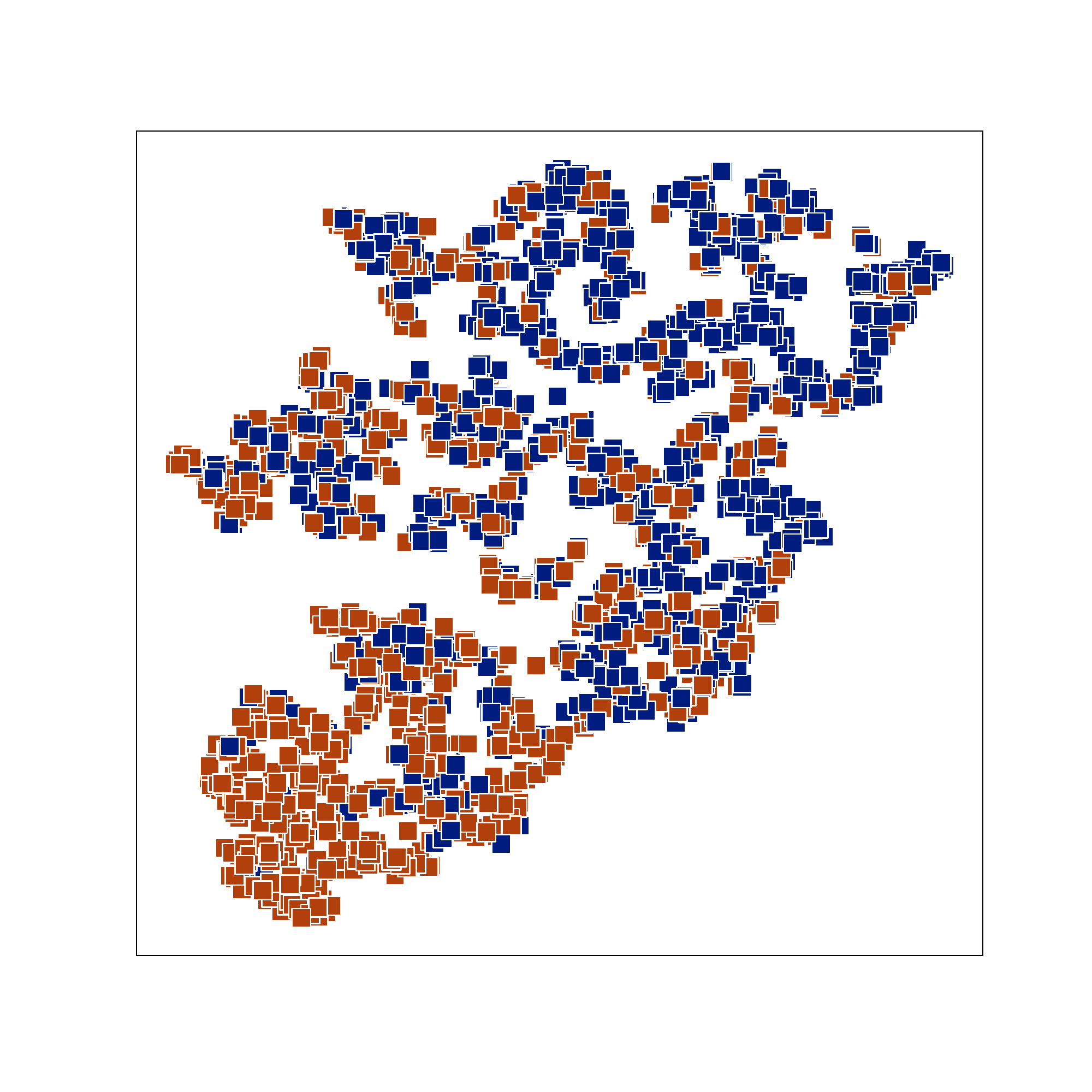}
    \end{minipage}
    \begin{minipage}[c]{0.15\linewidth}
    \centering{FC4}\\
    \includegraphics[width=\linewidth]{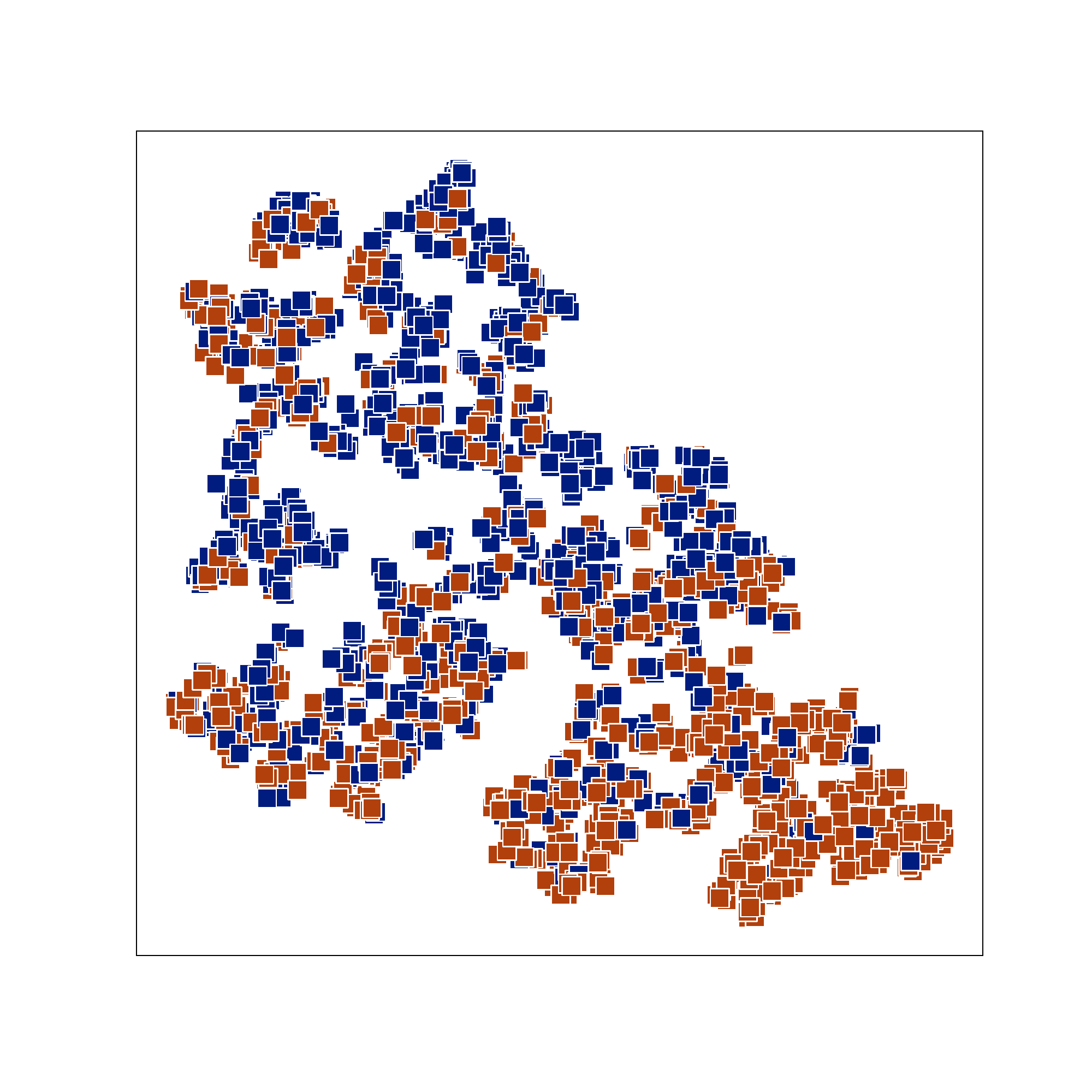}
    \end{minipage}
    \begin{minipage}[c]{0.15\linewidth}
    \centering{FC6}\\
    \includegraphics[width=\linewidth]{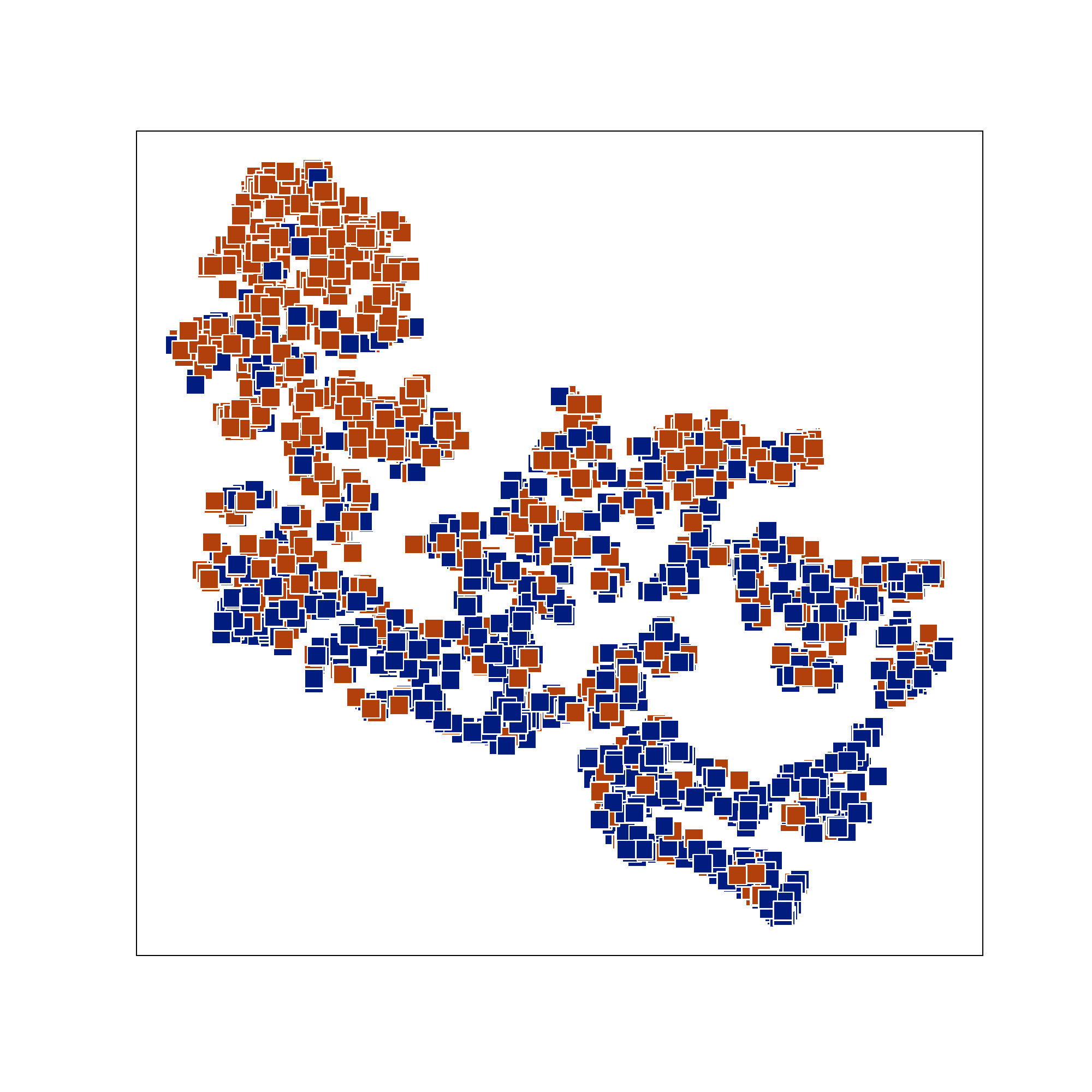}
    \end{minipage}
    \begin{minipage}[c]{0.15\linewidth}
    \centering{FC8}\\
    \includegraphics[width=\linewidth]{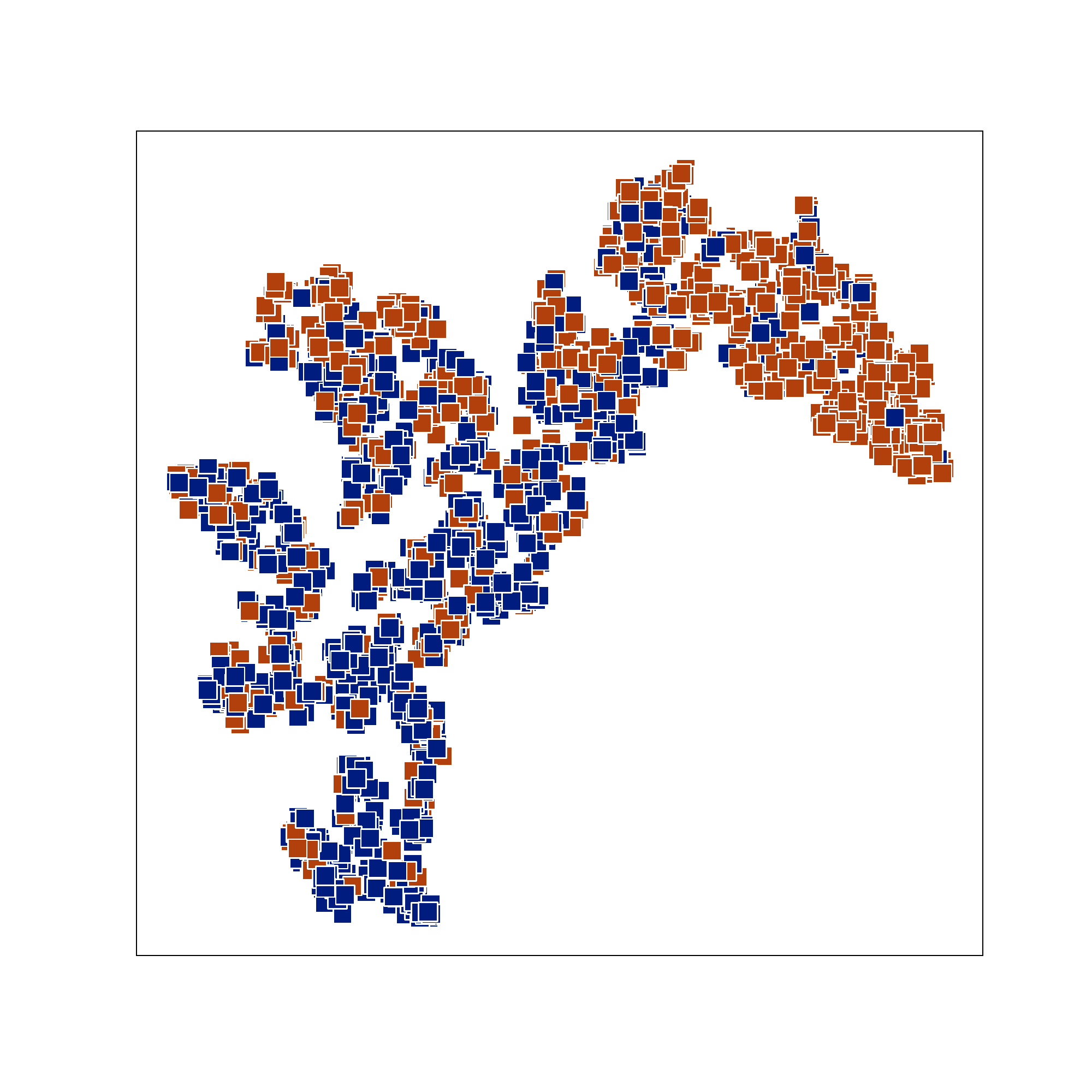}
    \end{minipage}
\begin{minipage}[c]{0.15\linewidth}
\centering{FC10}\\
    \includegraphics[width=\linewidth]{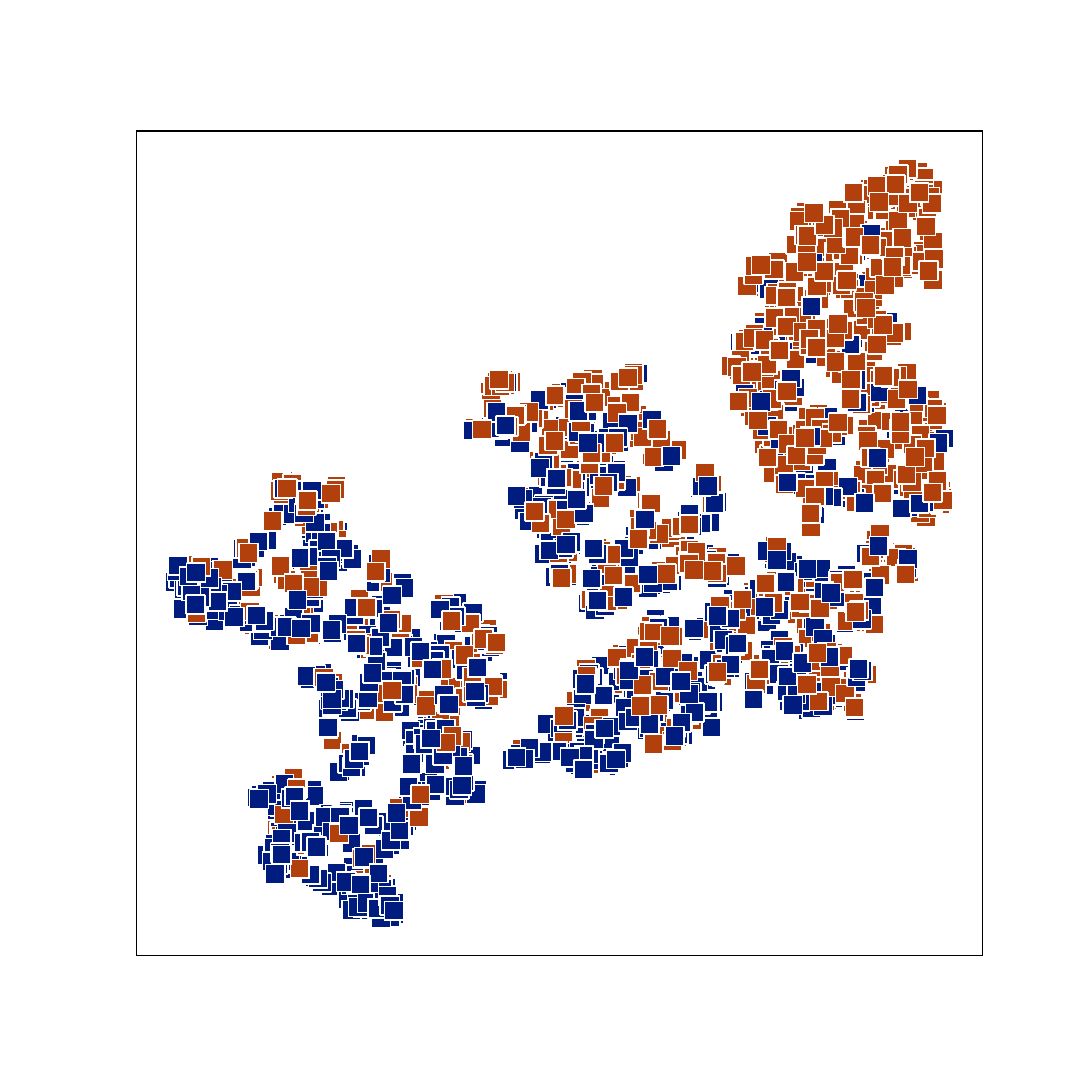}
    \end{minipage}
    \begin{minipage}[c]{\linewidth}
    \centering{c) Proposed approach without manifold regularization on test data.}\\
    \end{minipage}
    \begin{minipage}[c]{0.15\linewidth}
    \centering{Conv Section}\\
    \includegraphics[width=\linewidth]{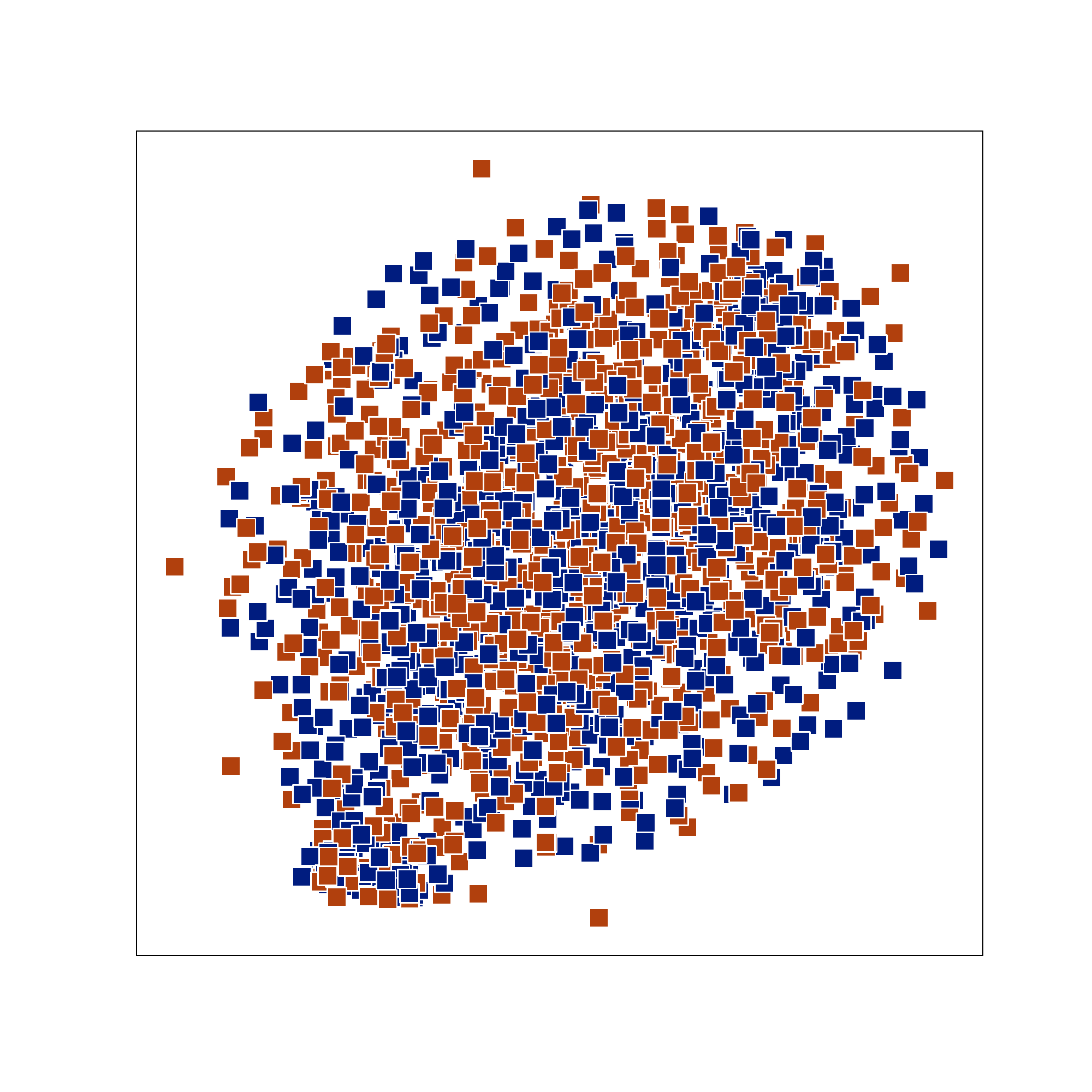}
    \end{minipage}
    \begin{minipage}[c]{0.15\linewidth}
    \centering{FC2}\\
    \includegraphics[width=\linewidth]{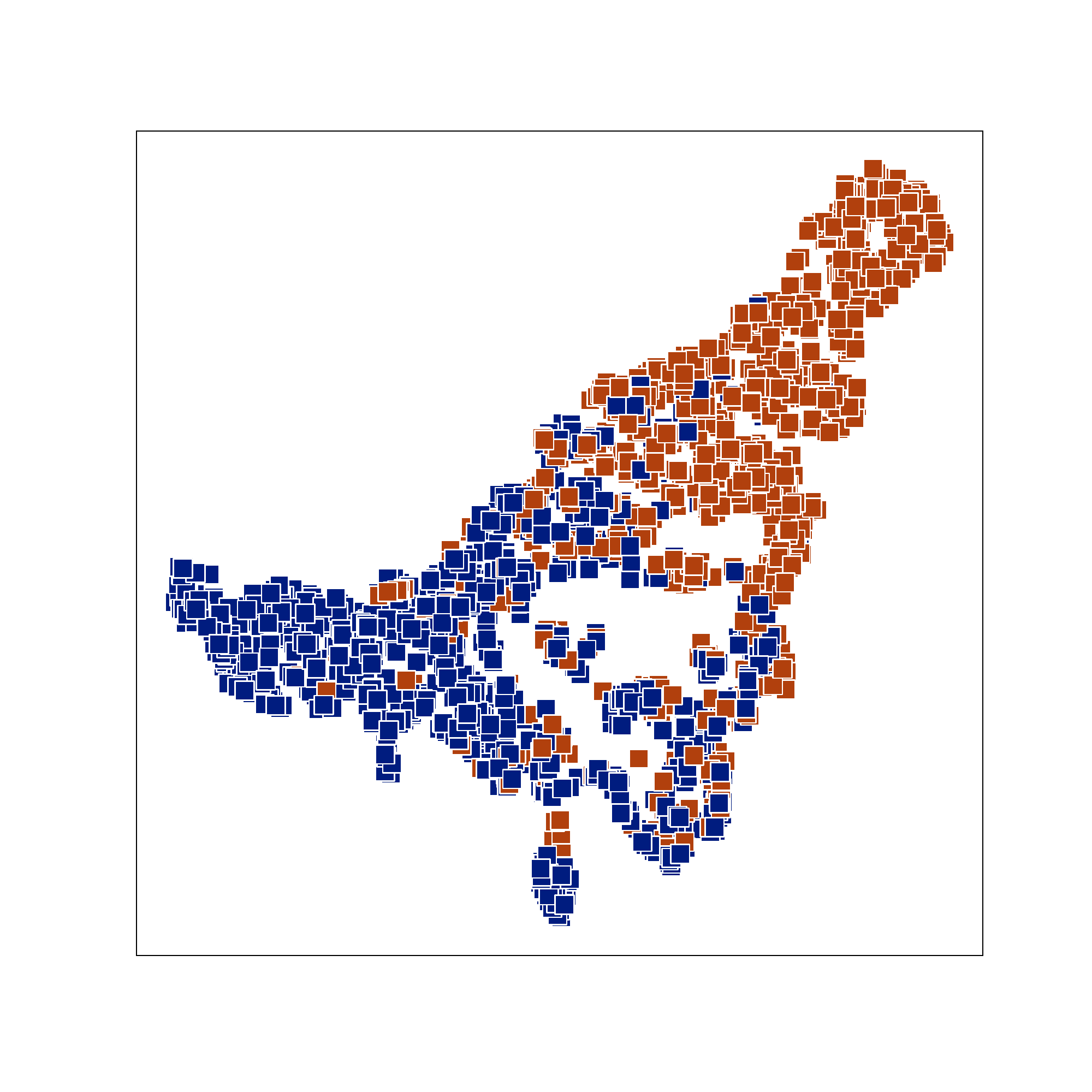}
    \end{minipage}
    \begin{minipage}[c]{0.15\linewidth}
    \centering{FC4}\\
    \includegraphics[width=\linewidth]{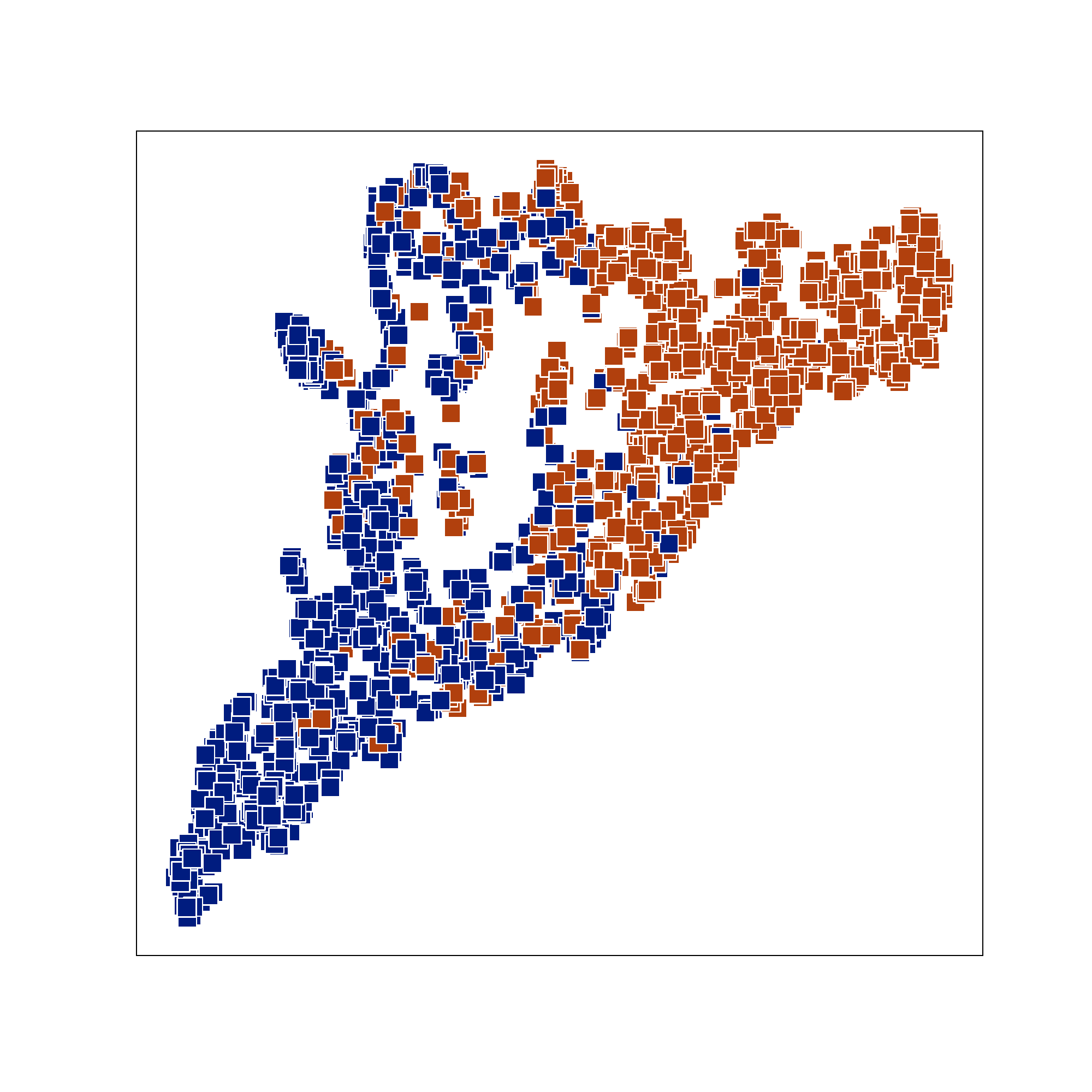}
    \end{minipage}
    \begin{minipage}[c]{0.15\linewidth}
    \centering{FC6}\\
    \includegraphics[width=\linewidth]{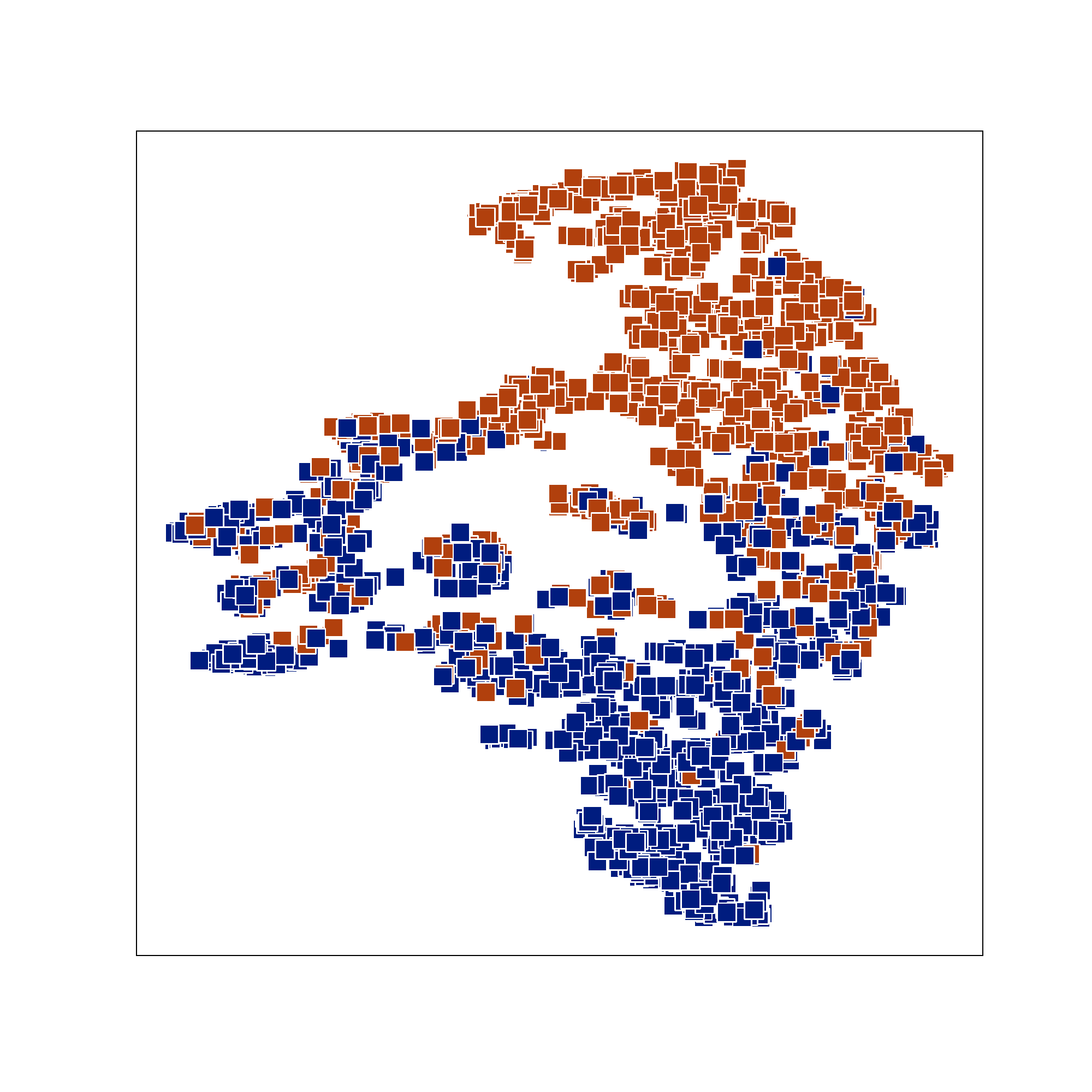}
    \end{minipage}
    \begin{minipage}[c]{0.15\linewidth}
    \centering{FC8}\\
    \includegraphics[width=\linewidth]{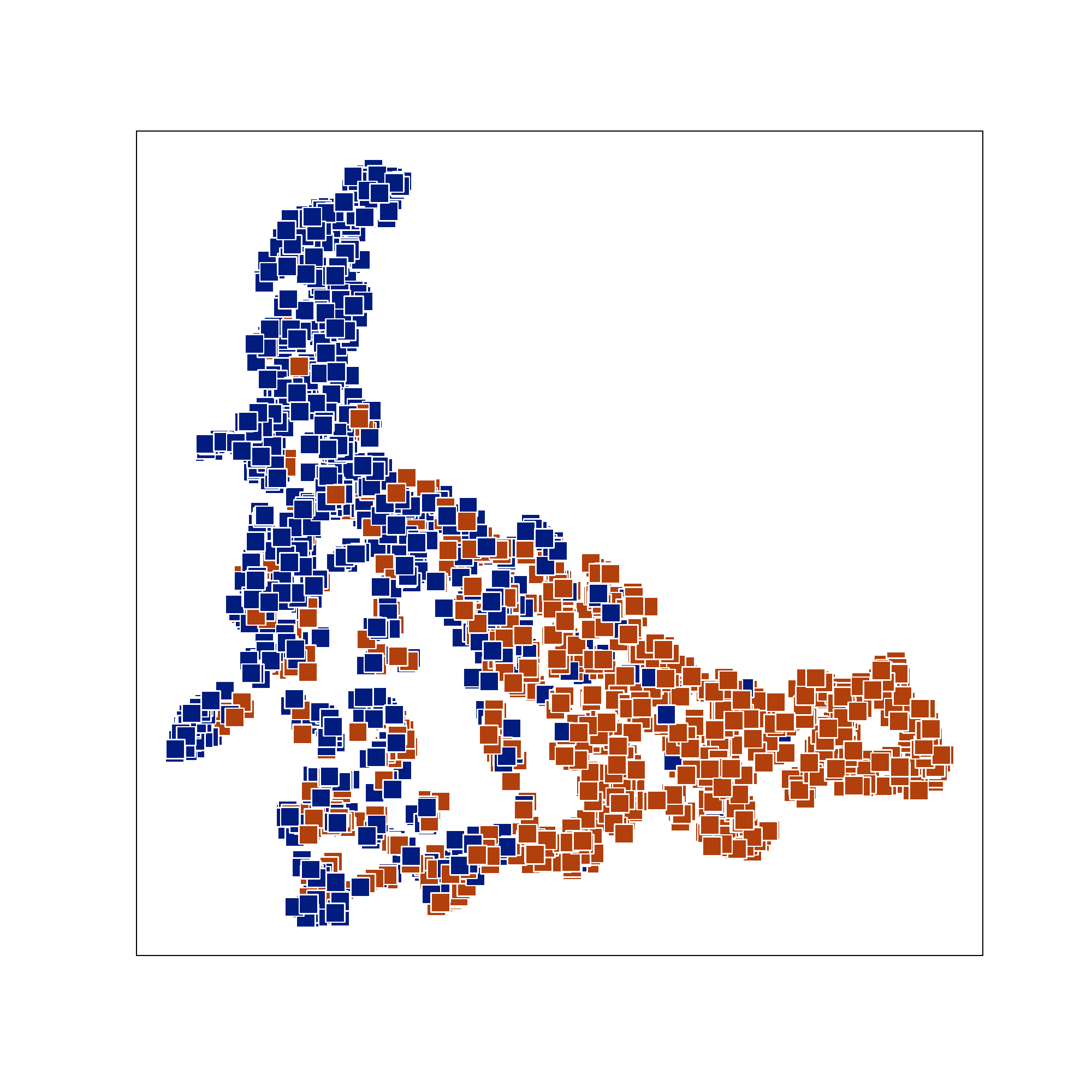}
    \end{minipage}
\begin{minipage}[c]{0.15\linewidth}
\centering{FC10}\\
    \includegraphics[width=\linewidth]{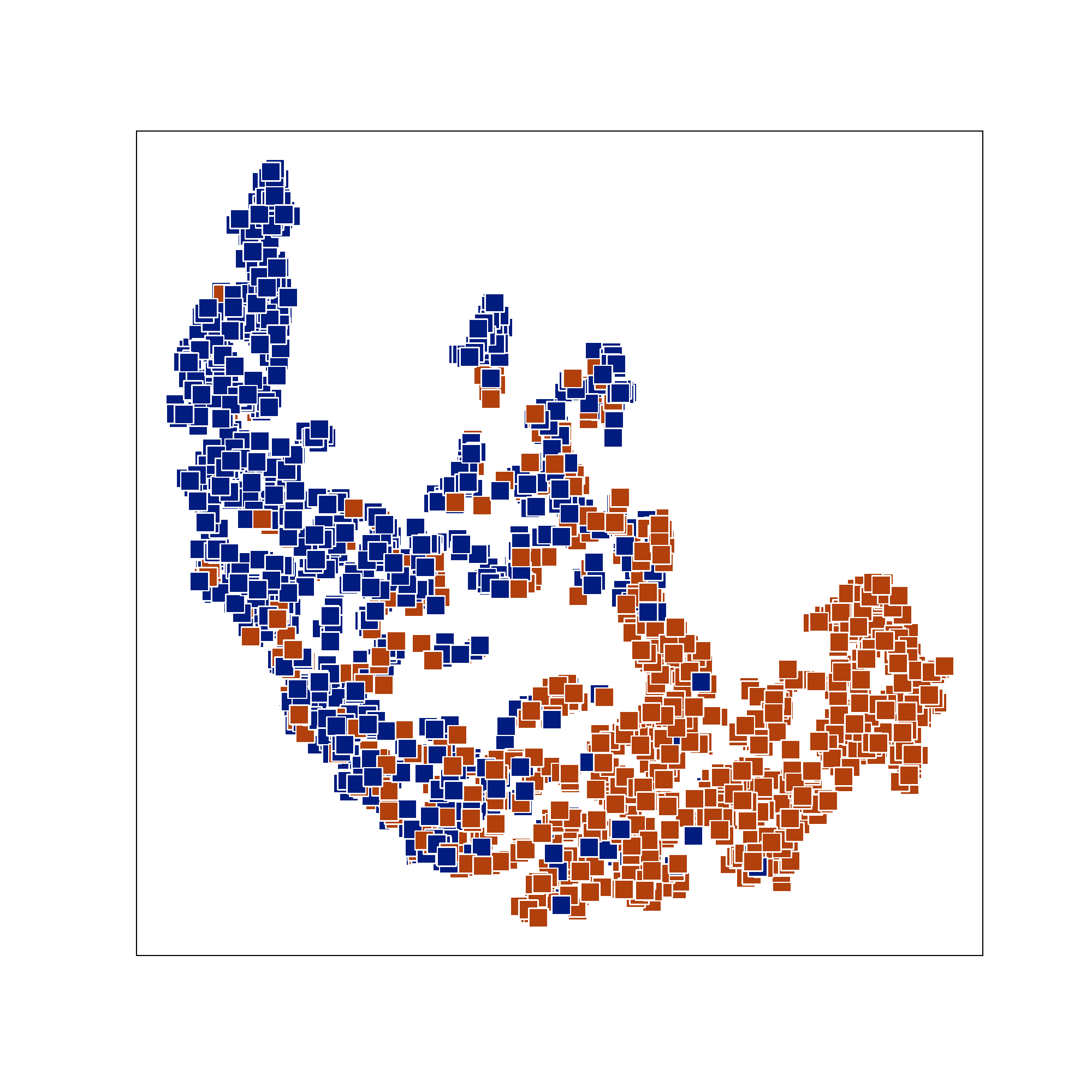}
    \end{minipage}
    \begin{minipage}[c]{\linewidth}
    \centering{d) Proposed approach with manifold regularization on test data.}\\
    \end{minipage}
    \caption{t-SNE plots of intermediate activations for data points in the ``dog" and ``cat" classes of the CIFAR-10 dataset. Figures (a) and (c) ((b) and (d)) illustrate the low-dimensional behavior of features produced without (with) manifold regularization.}
    \label{fig:tsne}
\end{figure*}

\subsection{Representation Learning Analysis}
In order to visualize the impact of the proposed framework on the learning process, we leverage \textit{t}-distributed stochastic neighbor embedding (t-SNE)\cite{10.5555/2968618.2968725} for dimensionality reduction of activations corresponding to the ``cat" and ``dog" classes in CIFAR10 at intermediate layers of the bottleneck portion of the network, with and without the manifold-oriented regularizer being enforced. The results are illustrated in Fig.~\ref{fig:tsne}: Figs.~\ref{fig:tsne}(a) and (b) ((c) and (d)) illustrate activations from samples in the training (test) set; Figs.~\ref{fig:tsne}(a) and (c) ((b) and (d)) contain visualizations of the activations without (with) manifold regularization; lastly, the first column shows the visualization of the activations at the end of the convolutional section of the network, and subsequent columns to the right include visualizations of activations as the data points move through the bottleneck section of the network. These results correspond to training with a mini-batch size of $5$, with $W=16$.

Note how, regardless of the use of a regularizer, the features produced by the convolutional section of the network showcase a high degree of overlap as well as highly isotropic distributions. Further downstream as the data enters the bottleneck section of the network, the low-dimensional visualizations seem to indicate that separation between the classes of interest for both the training and test is more evident when the regularization mechanism is enforced, in spite of it being unsupervised in nature, i.e., not leveraging labels. The visualizations further show that the method is effective at enforcing and preserving a manifold-like structure in the activations despite the mini-batch size being extremely small, which, as stated before, can lead to noisy gradient estimates and imperfect learning. 

\begin{figure}[t]
  \centering
   \includegraphics[width=0.9\columnwidth]{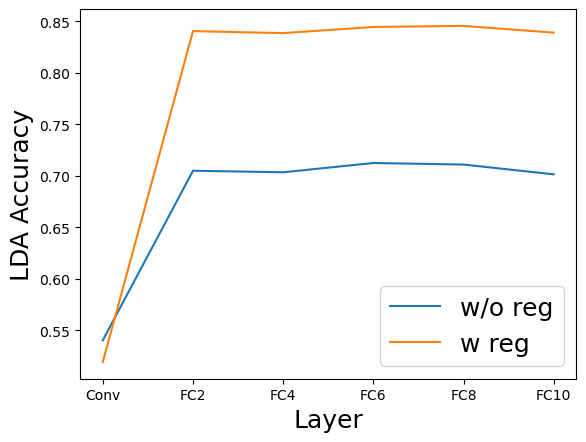}
   \caption{Discriminability of the activations at different stages of the network as measured by the accuracy of an LDA algorithm on the two-dimensional t-SNE feature representation of the data from Fig.~\ref{fig:tsne}, both with and without the manfold regularizer.}
   \label{fig:LDA}
\end{figure}

In order to quantify the degree of separability of the two classes in question based on the multiple intermediate representations, we performed linear discriminant analysis (LDA) on the two-dimensional t-SNE representation of features from the test set across the different layers and measured the performance (accuracy) of the discriminant. The results are shown in Fig.~\ref{fig:LDA}. It can be observed that the linear discriminant function performs almost at a chance level for both sets of features coming out of the convolutional section of the network. As soon as the data enters the bottleneck section of the network, the separation between the classes increases in both cases. However, it can be observed that the use of the manifold regularizer does a better job at increasing feature discriminability. This, in spite of the regularizer being unsupervised and having to rely on the extremely small set of samples in the mini-batch.


\section{Conclusions}
The memory resources that a given deep learning model consumes often determine the accessibility level of the associated technology, not only for the end-user but also for the community contributor. While \textit{post hoc} network compression techniques aid deployment of large pre-trained models are plentiful, methods that facilitate training of effective, high-capacity models are less common. In this paper, we introduced a framework that addresses this limitation by achieving significant memory savings \textit{during} model training. The effectiveness of the method at enabling training with compact networks and extremely small mini-batch sizes was demonstrated. The proposed framework acts in a manner that's independent to previously introduced efforts on network compression so it can be seamlessly combined with those to achieve further memory savings.

{\small
\bibliographystyle{ieee_fullname}
\bibliography{egbib}
}

\end{document}